**Application of Artificial Intelligence in Schizophrenia Rehabilitation Management: A Systematic Scoping Review**

Hongyi Yang, Zhao Liu*, Fangyuan Chang, Dian Zhu, Muroi Fumie

School of Design, Shanghai Jiao Tong University, Shanghai, China

[*] **Corresponding authors:** Zhao Liu
School of Design, Shanghai Jiao Tong University
800 Dongchuan Rd, Minhang District, Shanghai, China
Tel: +86 18801971294
Fax: +86 02134207231
Email: hotlz@sjtu.edu.cn

## Abstract

This systematic review assessed the current state and future prospects of artificial intelligence (AI) in schizophrenia rehabilitation management. We reviewed 61 studies on AI-related data types, feature engineering methods, algorithmic models, and evaluation metrics published from 2012 to the 2024. The review categorizes AI applications into the following key application areas: symptom monitoring, medication management, risk management, functional training, and psychosocial support. Findings indicate that supervised machine learning techniques, particularly for symptom monitoring and relapse risk management, remain the predominant approaches, effectively leveraging structured data while incorporating interpretable algorithms. This study underscores the potential of AI in transforming long-term management strategies for schizophrenia, offering valuable insights into improving the quality of life of patients. Future research should focus on expanding data sources through multimodal data integration, exploring deep learning models, and integrating AI-driven interventions into training tasks to fully capitalize on AI's potential in schizophrenia rehabilitation.

## Introduction

Schizophrenia is a severe and persistent mental disorder characterized by fundamental disturbances in cognition, perception, affect, and psychosocial functioning. The clinical manifestations typically include positive symptoms (hallucinations, delusions), negative symptoms (diminished emotional expression, avolition), disorganized behavior, and cognitive impairments, which collectively impair occupational and social functioning[1-3]. According to the World Health Organization (WHO), schizophrenia affects approximately 24 million individuals globally, with a lifetime prevalence of 0.32%[4-5].

The chronic nature and pervasive functional impairments associated with schizophrenia necessitate sustained, comprehensive rehabilitation interventions beyond acute symptom management. This long-term rehabilitation process requires continuous assessment and adaptation of interventions throughout the patient's recovery journey. The World Health Organization defines psychiatric rehabilitation as "a process that facilitates the opportunity for individuals with mental health disorders to reach their optimal level of functioning in the community, by both improving individuals' competencies and introducing environmental changes"[new ref]. Similarly, the Psychiatric Rehabilitation Association describes it as "services that facilitate the development of skills and access to supports necessary for success in living, working, learning, and social environments of choice"[new ref].

In recognition of these rehabilitation needs, the WHO's Comprehensive Mental Health Action Plan 2013-2030 advocates for a paradigm shift from institutional care to community-based psychiatric rehabilitation[6]. Community-based rehabilitation, characterized by integrated service delivery within patients' natural living environments, has emerged as the predominant model for psychiatric rehabilitation and management. This transformation aligns with contemporary psychiatric rehabilitation principles, emphasizing recovery-oriented, person-centered care that promotes autonomy and community integration.

Within this community-based framework, comprehensive rehabilitation management for individuals with schizophrenia encompasses a systematic, evidence-based approach that integrates clinical treatment with psychosocial interventions. The rehabilitation framework comprises several core components: comprehensive functional assessment, symptom monitoring, medication adherence management, and psychosocial interventions including cognitive remediation, psychoeducation, and social skills training[7-8]. These interventions are delivered within a recovery-oriented framework that emphasizes personal goal setting, skill development, and environmental supports[9]. The ultimate aim is to optimize psychosocial functioning, enhance quality of life, and facilitate community integration through individualized rehabilitation planning and implementation[10-12].

However, realizing this vision faces numerous challenges, including limited resources, pharmacological treatment side effects, poor patient compliance, and difficulties in maintaining long-term treatment effects[7, 13]. According to the WHO, only approximately 31.3% of patients receive specialized mental healthcare services[4]. A United States study indicated that 62% of adults with psychotic disorders and 41% of adults with severe mental illness had not received any mental healthcare in the past year [14], highlighting the critical issues of the global accessibility and quality of mental health services.

Thus, digital technologies have opened new avenues for enhancing and extending schizophrenia rehabilitation, not only complementing traditional clinical approaches, but also introducing innovative methods for delivering care [15]. Digital interventions, including mobile applications, virtual reality (VR) platforms, telemedicine services, and wearable devices, have demonstrated value in multiple facets of mental health management, offering more accessible and continuous interventions [16-18]. For instance, mobile applications can monitor emotional states and medication adherence while delivering psychoeducational content[19-21]; VR interventions simulate social scenarios to improve social skills and reduce anxiety [22-23]; telemedicine platforms facilitate access to professional mental health support, particularly in remote or underserved areas [24-25]; and wearable devices provide real-time physiological data for personalized treatment recommendations[26].

The integration of these digital technologies into mental healthcare has generated large-scale multimodal datasets that capture text, audio, video, physiological signals, and behavioral patterns from diverse contexts. These rich data sources enable researchers to systematically assess patient conditions and track rehabilitation progress over time [27-28]. By leveraging advanced artificial intelligence (AI) techniques, these datasets provide a foundation for developing predictive models and personalized interventions for schizophrenia rehabilitation [29-30].

AI refers to the science and engineering of creating systems or machines capable of performing tasks that typically require human intelligence such as perception, reasoning, and learning [31]. Within the AI domain, machine learning (ML) serves as the cornerstone of methodology. ML encompasses various statistical and algorithmic approaches that enable computers to identify patterns, dependencies, and structures within the data without being explicitly programmed[32]. Classic ML techniques, including linear [33] and Logistic Regression [34], decision trees [35], Random Forests (RF), and Support Vector Machines (SVM) [36], offer computationally efficient solutions to tasks such as classification, regression, and anomaly detection. More recent ensemble-based methods, such as gradient boosting frameworks [37] (e.g., XGBoost [38]), further enhance predictive accuracy. In addition to these supervised approaches, unsupervised methods (e.g., k-means [39], Gaussian mixture models [GMM] [40], principal component analysis [PCA] [41]) enable the discovery of latent structures, dimensionality reduction, and pattern recognition within unlabeled data, facilitating deeper insights into inherent groupings and feature interdependencies.

Deep learning (DL), a specialized subset of ML, leverages multilayered artificial neural networks to automatically learn complex hierarchical representations of data [42]. By stacking numerous hidden layers, often numbered in tens or hundreds, DL architectures can extract subtle patterns and high-level abstractions directly from raw inputs such as images, audio signals, or natural language text. Convolutional neural networks (CNNs), including landmark architectures, such as VGGNet [43] and ResNet [44], have been used to capture spatial and temporal features from visual and time-series data. Recurrent neural networks (RNNs) enhanced by variants, such as Long Short-Term Memory (LSTM) [45] and Gated Recurrent Unit (GRU) [46] units, have propelled advances in sequence processing, whereas transformer-based models [47] have redefined language modeling and representation learning. Word2Vec [48] and Bidirectional Encoder Representations from Transformers (BERT) [49] techniques have laid the foundation for more sophisticated architectures, culminating in the development of generative pre-trained transformers (GPT) [50] and large language models (LLMs), which excel at understanding and generating human-like language across diverse tasks [51-53].

Unlike ML methods, reinforcement learning (RL) focuses on sequential decision making in dynamic environments. RL agents learn optimal policies through trial-and-error interactions guided by a reward signal that provides feedback on the quality of the chosen actions [54]. Drawing on the principles of control theory and behavioral psychology, RL techniques enable systems to adaptively refine their strategies over time. The integration of deep learning within RL, referred to as deep reinforcement learning, combines the representation-learning power of DL with the decision-making capabilities of RL [55]. This integration allows for the efficient handling of high-dimensional states and action spaces, enabling the development of complex policies through end-to-end optimization. Notable algorithms, such as the Deep Deterministic Policy Gradient (DDPG) [56], Twin Delayed Deep Deterministic Policy Gradient (TD3) [57], and Batch-Constrained Deep Q-learning (BCQ) [58], exemplify how deep reinforcement learning extends the applicability of RL to challenging real-world scenarios.

These AI methods have demonstrated significant utility in elucidating the pathophysiology and improving diagnostic accuracy of schizophrenia. Advanced deep learning architectures have enabled automated analysis of neuroimaging data, identifying novel biomarkers through pattern recognition in structural and functional brain images. Machine learning algorithms have also enhanced diagnostic precision by integrating multimodal data, including neuroimaging, genetic markers, clinical symptoms, and cognitive assessments [59-63]. These technological advances not only contribute to our understanding of disease mechanisms but also offer promising tools for early detection and precision diagnostics.

The integration of AI into healthcare, including schizophrenia rehabilitation, introduces several complex challenges. These include ensuring the interpretability of algorithmic outputs, mitigating data bias, and maintaining robust patient safety standards [64-65]. Such considerations may affect the equitable and reliable implementation of AI-driven interventions, ultimately influencing patient outcomes and stakeholder trust. Thus, in-depth understanding of current AI applications in schizophrenia rehabilitation is crucial. By critically examining existing methodologies and evidence, future research can better integrate emerging AI technologies, particularly in the face of rapid technological advancements, while considering the limitations and implementation challenges that must be addressed before deployment. This critical approach ensures the responsible and effective use of AI in schizophrenia rehabilitation, guides informed decision making, and minimizes potential risks.

While existing systematic reviews examine AI applications in mental health, they inadequately address AI's role in schizophrenia rehabilitation. Current literature primarily focuses on AI in diagnosis and acute treatment [66-67], overlooking its applications in psychiatric rehabilitation management - a distinct phase encompassing functional assessment, psychosocial intervention delivery, and longitudinal outcome monitoring. Moreover, while some reviews explore AI in mental health rehabilitation broadly [68-69], they lack specific analysis of schizophrenia rehabilitation, where unique cognitive deficits and negative symptoms require specialized intervention approaches.

This systematic review addresses these limitations by examining AI applications specifically in schizophrenia rehabilitation management. We analyze technical implementations (algorithms, data modalities, feature engineering) and their practical applications in core rehabilitation domains: functional assessment, psychosocial intervention delivery, medication adherence monitoring, and community reintegration support. This focused examination provides insights into how AI technologies can enhance evidence-based psychiatric rehabilitation practices and improve longitudinal outcomes for individuals with schizophrenia.

**Table 1** Characteristics of studies using AI in schizophrenia rehabilitation management.

| Basic Information | | | | AI-Related Information | | | | Rehabilitation and Management Applications Areas | |
|---|---|---|---|---|---|---|---|---|---|
| **Authors and Year** | **Country/Region** | **Sample characteristics** | **Data Types** | **Data Preprocessing/Feature Engineering/Representation Learning** | **Model** | **Metric** | **Model Interpretability** | **Application Areas** | **Application Descriptions** |
| Howes et al. (2012)[70] | United Kingdom | 131 patients with schizophrenia. | Natural Language Data (conversation texts) | Dialogue Feature Extraction: Extract high-level dialogue features (e.g., number of turns, word count, number of questions) and word-level features (e.g., unigrams). Perform feature selection using the CfsSubsetEval selector. | SVM and Decision Tree | Acc: 91.1% (imbalanced) vs. 93.0% (balanced) | Feature Importance Analysis: Analyzed vocabulary in conversations (e.g., specific words related to symptoms or treatment) to identify important features associated with treatment adherence prediction. | Medication management | Using conversational text analysis to predict treatment adherence, enabling early identification of non-adherent patients and informing personalized interventions and adjustments to treatment plans. |
| Osipov et al. (2015)[71] | United Kingdom, United States | 16 patients with schizophrenia (final analysis included 12 patients after removing some | Physiological Data (heart rate, derived from ECG, sampled at 10-minute intervals) and Physical Activity | Statistical and Entropy Feature Extraction: Extract basic statistical features (mean, standard deviation) as well as multiscale entropy (MSE) | SVM (RBF kernel) | Acc: 95.3%; AUC: 0.99 | Not available. | Symptom monitoring and assessment | Demonstrating that physiological and behavioral signals can effectively classify schizophrenia, offering potential for |
| | | records due to data quality issues) and 19 healthy controls. The average disease duration was 6.4 years. Patients were in symptomatic remission and receiving antipsychotic medication. | Data (accelerometer, sampled at 5-minute intervals) | and transfer entropy (TE). | | | | | continuous monitoring and early relapse detection. |
| Wang et al. (2016)[72] | United States | 21 discharged patients with schizophrenia, with a follow-up period ranging from 64 to 254 days. | Sensor Data (behavior, location, audio, illumination, calls, etc.) collected via smartphone | Behavioral Feature Aggregation and Standardization: Aggregate daily behavioral features (walking time, dialogue frequency, sleep duration) and apply zero-mean and unit-variance standardization. | GBRT and RF | MAE: 1.378–2.29; r: 0.77–0.89 | Feature Importance Analysis: Conducted feature importance analysis using Random Forest (RF). | Symptom monitoring and assessment, Risk management | Leveraging smartphone sensor data to detect and forecast changes in mental health states (e.g., depression, stress, auditory hallucinations), facilitating real-time symptom monitoring and early clinical intervention. |
| Wang et al. (2017)[73] | United States | 36 outpatient patients with schizophrenia, in the post-discharge management | Behavioral Data (activity, calls, location), Sensor Data (sound, light), and Self-Reported Data (EMA | Smartphone Data Feature Extraction: Extract features from smartphone behavioral data and perform feature selection using linear | GBRT | Leave-One-Out Cross-Validation: MAE: 1.45; r: 0.70 | Feature Importance Analysis: Performed feature importance | Symptom monitoring and assessment | Predicting weekly symptom fluctuations (e.g., BPRS scores) using passive sensing data, enabling early identification of |
| | | phase, including a subgroup at high risk of relapse. | questionnaires) | regression and machine learning methods. | | | analysis using Gradient Boosted Regression Trees (GBRT) and analyzed the relationship between features and Brief Psychiatric Rating Scale (BPRS) scores. | | relapse risk, and informing timely clinical interventions and treatment adjustments. |
| Birnbaum et al. (2017)[74] | United States | 146 patients with schizophrenia. | Social Media Text Data (Twitter posts) | Language Feature Extraction: Utilize n-gram models and LIWC for linguistic analysis. | RF | Acc: 88%–95%; ROC AUC: 0.88–0.95; F1: 0.9 | Feature Importance Analysis: Conducted feature importance analysis using RF. | Symptom monitoring and assessment | Employing social media–based linguistic analysis for early psychosis detection, aiming to shorten the duration of untreated psychosis and support earlier therapeutic interventions. |
| Fond et al. (2019)[75] | France | 549 patients with schizophrenia (in a stable state). | Symptom Data (PANSS scores, aggression scores), Demographic Data, and Hospitalization Records | Not available. | Decision Tree (CART) | Relapse Prediction Acc: 63.8%; Sensitivity: 71%; Specificity: 44.8% | Feature Importance Analysis: Performed feature importance analysis using Decision Tree | Risk manageme nt | Using AI models to predict the two-year relapse risk and potential treatment disengagement in individuals with schizophrenia, thereby supporting proactive |
| | | | | | | | models. | | care planning. |
| Brodey et al. (2019)[76] | United States | 353 participants, including clinical high-risk (CHR) individuals and first-episode psychosis (FEP) patients, among whom 71 were low-risk controls (CLR). The study covered a 12-month conversion rate and a 24-month follow-up. | Self-Reported Data (Likert scales) | Spectral Clustering: Apply spectral clustering algorithms for data classification. | SVM (RBF kernel) | Positive Predictive Value (PPV): 76.5% (EPSI); 86.6% (EPSI+SIPS); Sensitivity: 51%; Specificity: 89% | Not available. | Symptom monitoring and assessment | Classifying clinical high-risk (CHR) individuals to evaluate imminent psychosis risk, guiding earlier and more targeted interventions. |
| Hulme et al. (2019)[77] | United Kingdom | 49 patients with schizophrenia. | Symptom-Related Data (anxiety, hallucinations, depression, etc.), collected via Ecological Momentary Assessment (EMA) | Multivariate Sequence Modeling: Represent data as multivariate sequences, though specific feature engineering methods are not described. | Hidden Markov Models (HMMs) | Bayesian Information Criterion (BIC): Optimal = 44,929 | Model Interpretability : Explained the model through hidden states and transition probabilities, describing psychological phenotypes of different states. | Symptom monitoring and assessment | Applying models that identify latent symptom subgroups and predict high-risk clinical events (e.g., relapse) to enable real-time, tailored care strategies. |
| Cohen et al. (2020)[78] | United States | 121 patients with severe mental illness, | Audio Data | Voice Feature Extraction (GeMAPS, CANS): Extract voice features | Lasso Regression | Classification Acc: 90%–99% | Feature Selection and Importance | Symptom monitoring and | Utilizing digital phenotyping to track symptom severity, |
| | | including 76 patients with schizophrenia and others with depression and bipolar disorder. The average age was 41.88 years. | | (frequency, intensity, pitch) using the GeMAPS and CANS feature sets. | | | Analysis: Identified key features using stability selection analysis and conducted subsequent analyses. | assessment | enhance diagnostic accuracy, evaluate treatment efficacy, and anticipate relapse or social functioning challenges. |
| Wu et al. (2020)[79] | Taiwan | 32,277 first-episode patients with schizophrenia (average age 36.7 years; 48.8% male). | Demographic Data, Baseline Clinical Data, and Medication Data | Not available. | Ensemble Machine Learning (Super Learner) | Treatment Success Rate (ITR-based): 51.7% | Feature Importance Analysis: Conducted feature importance analysis using RF. | Medication management, Risk management | Forecasting individual treatment success rates to optimize medication strategies, reduce hospitalization rates, and improve overall rehabilitation outcomes. |
| Wang et al.,(2020)[80] | Canada | 275 patients diagnosed with schizophrenia (age range 18–75 years). | Clinical Data, Demographic Data, and Sociocultural Variables | Not available. | RF, LR, SVM, among seven classification algorithms | RF Performance: Highest Acc: 62%; AUROC: 0.63 | Not available. | Risk management | Predicting patients' propensity toward violent behavior to guide early clinical decision-making and reduce associated risks in a rehabilitation context. |
| Birnbaum et al., (2020)[81] | United States | 116 participants, including 42 patients with schizophrenia, schizotypal disorder, and | Internet Search Data (timestamped query records) | Lexical Classification and Temporal Distribution: Use LIWC for lexical classification and extract features like query frequency and temporal | RF, SVM, and Gradient Boosting (GB) | AUC: 0.74 (diagnostic, best: RF); 0.71 (relapse, best: SVM) | Feature Importance Analysis: Performed feature importance | Risk management, Symptom monitoring and | Analyzing patterns in patients' online search behaviors to predict relapse risk and symptom changes, providing digital tools |
| | | other schizophrenia spectrum disorders (SSD), and 74 healthy volunteers. | | length/distribution. | | | analysis using RF. | assessment | for earlier diagnosis and proactive intervention. |
| Arevian et al. (2020)[82] | United States | 47 patients with severe mental illness, of which 28% were patients with schizophrenia. | Audio Data | Semantic and Acoustic Feature Extraction: Extract semantic features (emotional vocabulary, lexical complexity) and acoustic features (fundamental frequency, harmonicity). | SVM | Spearman's Rho: 0.78 | Feature Correlation Analysis: Evaluated the correlation between each feature and provider ratings using Spearman correlation analysis. | Symptom monitoring and assessment | Examining semantic and acoustic speech features to track changes in clinical status over time, supporting continuous symptom monitoring and individualized rehabilitation plans. |
| Tseng et al. (2020)[83] | United States | 61 patients diagnosed with schizophrenia, with study data covering 6,132 days. | Behavioral Data (mobile sensor data, phone usage), Environmental Data (ambient light, sound), and Physiological Data (sleep duration) | Rhythmic Feature Extraction: Extract ultradian, circadian, and long-period rhythmic features. | Multi-Task Learning (MTL) and Multi-output Support Vector Regression (m-SVR) | RMSE: 0.309 | Feature Importance Analysis: Applied LASSO regularization to analyze feature importance. | Symptom monitoring and assessment, Risk management | Utilizing symptom prediction models and rhythmic behavioral/environmental indicators to guide early interventions, such as adjusting environmental conditions or activity patterns. |
| Wang et al.(2020)[84] | United States | 55 outpatient patients with schizophrenia. | Behavioral Data (activity frequency, communication | PCA and Behavioral Feature Extraction: Apply PCA to extract behavioral | RF, Extremely Randomized Trees (ET), and | Best model (Employment/Occupational): r: 0.60; MAE: 2.35 | Feature Correlation Analysis: Used | Symptom monitoring and | Predicting aspects of social functioning (e.g., social engagement, |
| | | | patterns, location visitation patterns), passively collected via smartphone | features and integrate them with machine learning models. | XGBoost | | PCA and GEE to extract behavioral patterns and establish associations between variables. | assessment, Psychosocial support or intervention | occupational capacity) through AI-based mobile data analysis, offering behavioral insights and informing intervention strategies. |
| Wang et al. (2021)[85] | Singapore | 22 patients with schizophrenia, recently discharged from a psychiatric hospital. | Digital Phenotype Data (behavioral data, wearable device data such as sleep, steps, heart rate; smartphone sensor data such as location information, light intensity, communication patterns) | Sensor Data Processing and Feature Construction: Clean sensor data and build time-series features (e.g., sleep duration, step count, heart rate variability). | Machine Learning Models (not explicitly stated) | Performance Metrics: Not available | Not available. | Risk management, Symptom monitoring and assessment | Remote collection and analysis of daily behavioral data to anticipate relapses, enabling earlier and more personalized interventions and the detection of potential clinical events. |
| Banerjee et al. (2021)[86] | United Kingdom | 1,706 patients with schizophrenia. | Electronic Health Records (EHR) and Biopsychosocial Factors Data (medications, comorbidities, social support, etc.) | Dimensionality Reduction and Encoding: Apply one-hot encoding and autoencoders for feature dimensionality reduction. | Autoencoder, RF, and LR | AUC: 0.80 (RF; 95% CI: [0.78, 0.82]) | Visual Explanation: Assisted in interpreting prediction results through class contrast analysis and heatmap generation. | Medication management, Risk management | Identifying mortality-related factors (e.g., medication effects, substance misuse) to inform personalized early interventions and integrated management of comorbid conditions, thus improving patient |
| | | | | | | | | | outcomes. |
| De Boer et al. (2023)[87] | The Netherlands | 142 patients with schizophrenia spectrum disorders and 142 healthy control subjects. | Audio Data | Voice Feature Extraction (OpenSMILE): Use OpenSMILE to extract voice features. | RF | Schizophrenia vs. Controls: Acc: 86.2%, AUC: 0.92; Positive vs. Negative Symptoms: Acc: 74.2%, AUC: 0.76 | Feature Importance Analysis: Conducted feature importance analysis using RF. | Symptom monitoring and assessment | Using voice parameters to identify schizophrenia patients and their symptom profiles, supporting symptom monitoring and evaluation in rehabilitation settings. |
| Ziv et al. (2022)[88] | Israel | 24 patients with schizophrenia (male, aged 19–63 years). | Natural Language Data (transcribed texts) | Morphological Feature Extraction: Extract morphological features (e.g., part-of-speech tagging, token-to-word ratio, type-to-token ratio). | RF, SVM, and XGBoost | Optimal Model Acc: 80%–90% | Not available. | Symptom monitoring and assessment | Distinguishing schizophrenia patients from healthy controls via linguistic feature analysis, demonstrating potential applications in ongoing symptom monitoring. |
| Lin et al., (2021)[89] | Taiwan | 302 diagnosed patients with schizophrenia (aged 18–65 years, with good physical health). | Clinical Symptom Data (SANS20, HAMD17, PANSS-Positive) and Cognitive Function Data (e.g., WAIS-III, social cognition, etc.) | Feature Selection (M5 Prime): Use the M5 Prime feature selection algorithm. | Bagging Ensemble, SVM, RF, and Linear Regression | Best Model RMSE (QLS Prediction): 6.4293 ± 1.1332 | Feature Importance Analysis: Identified important features using the M5 Prime feature selection algorithm through its recursive selection | Functional training, Symptom monitoring and assessment | Predicting functional outcomes in patients and assessing how negative, depressive, and positive symptoms influence their functional status, guiding targeted rehabilitation efforts. |
| | | | | | | | process. | | |
| Jeon et al. , (2022)[90] | South Korea | 45 patients with stable-phase schizophrenia. | Video Data (eye-tracking, facial expression emotion analysis) | Deep Learning Model Utilization: Employ a pre-trained deep learning model (Residual Masking Network, FER2013 dataset). | Residual Masking Network (RMN) | Acc: >74% | Not available. | Symptom monitoring and assessment, Functional training | Linking gaze avoidance patterns to psychopathological features and social interaction difficulties in schizophrenia, informing AI-based social skills training interventions. |
| Parola et al. (2021)[91] | Italy | 32 patients with stable-phase schizophrenia and an additional 35 healthy controls. | Natural Language Data, Non-Verbal Data (gestures, facial expressions, etc.), and Audio Data | Feature Evaluation: Assess feature information gain using decision tree models. | Decision Tree | Acc: 82%; Sensitivity: 76%; Specificity: 90%; Prec: 91%; AUC: 0.894 | Feature Importance Analysis: Performed feature importance analysis using Decision Trees. | Functional training, Symptom monitoring and assessment | Identifying pragmatic language deficits (e.g., sarcasm comprehension) to inform the design of targeted language-based training and rehabilitation interventions that improve social communication. |
| Rashid et al., (2021)[92] | Singapore | 100 patients with schizophrenia spectrum disorders (within 8 weeks post-discharge). | Sensor Data (GPS, accelerometer), Behavioral Data (call and SMS records), and Physiological Data (heart rate, sleep) | Not available. | Not available | Performance Metrics: Not available | Visual Explanation: Assisted in interpreting prediction results through class contrast analysis and | Risk manageme nt, Symptom monitoring and assessment | Predicting relapse from digital signal changes, offering early warnings and optimizing clinical interventions for sustained patient stability. |
| | | | | | | | heatmap generation. | | |
| Podichetty et al., (2021)[93] | United States | 639 diagnosed patients with schizophrenia. | Symptom Scores (PANSS) and Clinical Diagnosis Data | Correlation Analysis: Select features based on Pearson correlation. | RF | AUC: 0.7; Sensitivity (TPR): 0.194; Specificity (TNR): 0.936 | Feature Importance Analysis: Conducted feature importance analysis using RF. | Risk manageme nt | Identifying patients likely to be non-responsive to treatment, thereby increasing the proportion of effectively treated patients and improving rehabilitation planning. |
| Soldatos et al., (2022)[94] | Greece, Denmark | 280 first-episode patients with schizophrenia (179 from Greece and 101 from Denmark). | Clinical Scale Data (PANSS, HAM-D, GAF, PSP) | Variable Selection (Elastic Net): Employ the Elastic Net method for predictor selection. | SVM | AUC: 0.71 | Feature Selection and Impact Analysis: Selected important features using the Elastic Net method and reported feature impact values. | Symptom monitoring and assessment | Predicting early remission status in patients, aiding in identifying high-risk individuals who may not achieve remission, and informing functional assessments and social reintegration efforts. |
| Badal et al., (2021)[95] | United States | 218 patients with schizophrenia and 154 healthy controls. | Facial Emotion Recognition Data (reaction time, accuracy, confidence scores) | GINI Index Feature Selection: Use the GINI index for feature selection. | Neural Networks (ReLU, Tanh), RF, and SVM | Cohen's d: 0.87–1.24; AUC: 0.73–0.81; F1: 0.71–0.78 | Feature Contribution Analysis: Performed feature importance analysis using | Symptom monitoring and assessment | Predicting emotional response differences in schizophrenia patients, providing insights into social cognition and emotional processing |
| | | | | | | | RF. | | that can guide rehabilitation management. |
| Depp et al., (2022)[96] | United States | 86 participants, including patients with schizophrenia, schizoaffective disorder, and bipolar disorder with psychotic features, aged 19–65 years. | Image Data (facial emotion recognition tasks) | Not available. | Machine Learning Models not explicitly used (emotion recognition tasks) | Acc: ~75.6% | Not available. | Symptom monitoring and assessment | Using real-time facial emotion recognition to assess social cognitive abilities and their relation to hallucinations, distrust, and mood fluctuations, enhancing symptom monitoring with a social cognition dimension. |
| Narkhede et al., (2022)[97] | United States | 52 patients with mental illness (including 22 with schizophrenia) and 55 healthy controls. | Active Data (EMA surveys) and Passive Data (GPS location, accelerometer) | Feature Selection (RFE, Boruta): Apply RFE and the Boruta algorithm to select important features. | RF, L1 Regularization, H2O AutoML | Prec: 79%–91%; AUC: ≤0.864 | Feature Importance Analysis: Conducted feature importance analysis using RF. | Symptom monitoring and assessment | Employing digital phenotyping (e.g., geolocation, EMA surveys) to predict and monitor changes in negative symptoms, supporting early intervention and personalized rehabilitation strategies. |
| Kalinich et al., (2022)[98] | United States | 25 patients with schizophrenia and 30 healthy controls. | Smartphone Gamified Cognitive Task Metadata (interaction time, number of clicks, | Static and Time-Series Feature Extraction: Extract static and time-series features (e.g., total interaction time, | RF | AUC: 0.95 (mental illness prediction); 0.80 (sleep function prediction) | Feature Importance Analysis: Assessed feature | Symptom monitoring and assessment | Utilizing smartphone-derived data to predict disease states and sleep-related cognitive |
| | | | idle time, etc.) | click-time mean/variance). | | | contributions using SHAP. | | function, informing early interventions and patient-centered rehabilitation planning. |
| Price et al., (2022)[99] | United States | 38 patients with schizophrenia spectrum disorders (average age 31.42 years; 71.1% male). | Passive Phone Usage Data, Symptom Scores (BSI), and Semi-Structured Interview Data | Normalization: Normalize data (mean=0, std=1). | Ensemble Learning (Linear Models, Tree Models, Multilayer Perceptrons) | r: 0.32 (symptom), 0.39 (engagement), 0.34 (user feedback); $R^2$: 0.10–0.16 | Feature Importance Analysis: Evaluated feature contributions using SHAP. | Symptom monitoring and assessment, Psychosocial support or intervention | Predicting patient response and engagement with digital interventions based on initial symptoms and usage patterns, helping optimize rehabilitation measures and enhance patient experience. |
| Adler et al., (2022)[100] | United States | CrossCheck dataset (61 patients with schizophrenia, average age 37.11 years, 41% female, with a history of major psychosis-related events). | Behavioral Data (walking time, sleep patterns, phone unlock counts, etc.), Environmental Data (illumination, noise levels) | Not available. | GBRT | Evaluation Metrics: MAE, Sensitivity, Specificity, etc. | Not available. | Symptom monitoring and assessment | Using passive data to predict sleep quality and stress levels, supporting timely interventions and continuous symptom monitoring in schizophrenia rehabilitation. |
| Jeon et al. (2022)[101] | South Korea | 8,502 patients with early-onset schizophrenia (aged 2–18 years) receiving risperidone | Health Insurance Data (diagnosis records, prescription records, medical visit records) | Normalization and Feature Filtering: Apply min-max normalization and filter out low-frequency features. | RF, GBM, SuperLearner, Lasso Regression | AUC (Risperidone Adherence, GBM): 0.686 | Feature Importance Interpretation: Conducted feature importance | Medication management | Forecasting antipsychotic medication adherence to inform personalized treatment decisions and improve long-term |
| | | (N=5,109) or aripiprazole (N=3,393). | | | | | analysis using RF. | | treatment continuity. |
| Zhu et al. (2022)[102] | China | 393 patients with schizophrenia receiving olanzapine treatment. | Electronic Health Records (EHR) (biochemical indicators, polypharmacy information, drug concentration data) | Missing Value Imputation and Encoding: Impute missing values using the k-nearest neighbors algorithm, apply min-max normalization, and perform one-hot encoding. | XGBoost | MAE: 0.046; MSE: 0.0036; RMSE: 0.060; MRE: 11% | Feature Importance Analysis: Assessed feature contributions using SHAP. | Medication manageme nt, Symptom monitoring and assessment | Predicting changes in prolactin levels under certain treatments (e.g., olanzapine), monitoring side-effect risks, and informing safer, more effective medication strategies. |
| Richter et al. (2022)[103] | United States | 24 patients with schizophrenia and 20 healthy controls. The average age of patients was 41 years and of controls was 42 years. | Natural Language Data (intonation, jitter, speech rate) and Facial Dynamic Feature Data (3D facial landmarks extracted via Mediapipe) | Language and Facial Feature Extraction: Use Praat for linguistic acoustic features and Mediapipe for facial dynamics. | RF | AUC: 0.84 ± 0.02 (language), 0.75 ± 0.04 (facial), 0.86 ± 0.02 (fusion) | Feature Importance Analysis: Performed feature importance analysis using RF. | Symptom monitoring and assessment | Identifying schizophrenia patients through linguistic and facial feature analysis, enabling remote, personalized symptom evaluation and long-term monitoring. |
| Chen et al. (2023)[104] | Taiwan | 105 patients with schizophrenia, aged 20–65 years, in a stable phase at a day care center, with 94 completing the study. | Medication Adherence Data collected via smartphone cameras (face and medication appearance recognition) | Not available. | Model type not explicitly described (facial/behavio ral recognition) | Medication Adherence: 94.72% (experimental), 64.43% (control) | Not available. | Medication manageme nt, Symptom monitoring and assessment | Employing mobile applications to improve medication adherence, stabilize psychiatric symptoms, and enhance cognitive function, providing effective rehabilitation tools, particularly during periods of restricted face-to-face |

| | | | | | | | | | care. |
|---|---|---|---|---|---|---|---|---|---|
| Hudon et al. (2023)[105] | Canada | 18 patients with treatment-resistant schizophrenia, unresponsive to two or more antipsychotic medications. | Natural Language Data (transcriptions of therapy sessions) | Text Vectorization and Dimensionality Reduction: Vectorize text using TF-IDF and reduce dimensions with PCA. | K-Means Clustering | Elbow Method for Clustering: Metrics not reported | Not available. | Symptom monitoring and assessment | Analyzing patient-language interactions with virtual avatars to inform understanding of treatment-resistant schizophrenia symptoms and guide individualized rehabilitation approaches. |
| Miley et al., (2023)[106] | United States | 463 patients with early intervention schizophrenia. | Social Function Data, Occupational Function Data, Motivation Data, and Social Cognition Ability Data | Not available. | Greedy Fast Causal Inference (GFCI) model | Comparative Fit Index (CFI): 0.884; RMSEA: 0.066 | Causal Path Explanation: Achieved path explanation based on causal inference. | Functional training | Identifying causal pathways affecting social and occupational functioning, informing early intervention priorities and treatment target selection. |
| Wong et al., (2024)[107] | Hong Kong | 1,398 first-episode patients with schizophrenia, of whom 191 used clozapine, with a median age at first diagnosis of 20 years and a follow-up period | Clinical Record Data (baseline and 36-month follow-up data) | AutoML for Feature Selection: Utilize AutoML to streamline feature selection. | Automated Machine Learning (AutoML) and optimized tree models | AUC (36-Month): 0.774; Brier Score: 0.0906 | Feature Importance Analysis: Evaluated feature importance using tree-based models. | Risk management, Medication management | Using risk calculators to predict treatment resistance, guiding individualized intervention planning and evidence-based decisions regarding clozapine use. |
| | | of 12–17 years. | | | | | | | |
| Difrancesco et al., (2016)[108] | Italy, United Kingdom | 5 patients diagnosed with schizophrenia. | GPS Data (geolocation, speed, time, etc.) | Geolocation Clustering: Extract features using time- and density-based geolocation clustering algorithms. | K-Means Clustering (time- and density-based) | Rec: 0.771; Prec: 0.954 | Not available. | Symptom monitoring and assessment | Monitoring social functioning via GPS data, detecting changes in outdoor activities, and indirectly signaling relapse risk for timely preventive measures. |
| Amoretti et al. (2021)[109] | Spain | 144 non-affective psychosis patients (including those with schizophrenia), aged 18–35 years, with first-episode and a two-year follow-up. | Clinical Symptom Data (PANSS, MADRS scale scores) | PCA for Dimensionality Reduction: Apply PCA to reduce the number of features. | Fuzzy Clustering | Clustering: Dunn, Gamma, Silhouette >0.7; Discriminant Function Analysis (DFA): 93.1% (baseline), 94% (follow-up) | Not available. | Risk manageme nt, Symptom monitoring and assessment | Using clustering analysis to distinguish patient subgroups and symptom trajectories (e.g., good prognosis vs. clinical deterioration), aiding in disease progression prediction and tailored interventions. |
| Rodríguez-R uiz et al., (2023)[110] | Mexico | 22 patients with schizophrenia and 22 healthy control subjects (Psykose dataset). | Nighttime Physical Activity Data (accelerometer signals) | Statistical Feature Extraction and RFE: Extract 23 statistical features (mean, variance, skewness, kurtosis, etc.) and select the optimal subset via RFE. | RF | Nighttime Data Model Acc: 98.24% | Not available. | Symptom monitoring and assessment | Differentiating between schizophrenia, depressive disorders, and healthy controls based on nighttime activity patterns, potentially informing sleep-related interventions and |
| | | | | | | | | | evaluating treatment efficacy. |
| Zhou et al., (2022)[111] | United States | 63 patients with schizophrenia, average age 37.2 years, with monitoring duration exceeding 12 months. | Multimodal Sensor Data (ambient light, sound, conversations, accelerometer, etc.) | Behavioral Pattern Clustering and Feature Extraction (GMM, DTW): Use GMM and DTW to cluster behavioral patterns and derive personalized features. | Gaussian Mixture Model (GMM), Dynamic Time Warping (DTW), Balanced Random Forest (BRF) | Relapse Prediction F2 Score: 0.23 vs. baseline 0.042 | Feature Importance Evaluation: Assessed feature importance based on feature selection frequency in cross-validatio n. | Risk manageme nt, Behavior monitoring | Employing mobile sensing data to cluster behavioral patterns and predict relapse risk, enabling early detection of concerning behavioral changes and timely intervention. |
| Holmlund et al., (2020)[112] | United States | 104 participants, including 16 diagnosed patients with schizophrenia. | Audio Data (recorded by devices and transcribed into text) | NLP Methods (WMD, Semantic Vectors): Apply WMD and semantic vectors for text semantic analysis and similarity measurement. | Linear Regression and a custom Automatic Speech Recognition (ASR) system | Prediction vs. Manual Assessment r: 0.83 | Feature Importance Evaluation: Conducted feature importance analysis using Linear Regression models. | Symptom monitoring and assessment | Implementing voice-based frameworks for daily language memory tests and remote clinical monitoring, potentially tracking cognitive fluctuations and treatment responses. |
| Xu et al., (2022)[113] | United States | 384 patients with schizophrenia. | Natural Language Data (transcriptions of voice diaries) | Voice Coherence Analysis (TARDIS): Analyze voice coherence using time-series feature extraction (TARDIS). | Deep Learning Models (e.g., BERT, FastText) + Support Vector Regression (SVR) | ROC AUC (TARDIS): 0.744; Spearman's Rho: 0.396 | Not available. | Symptom monitoring and assessment | Using AI-driven voice analysis to quantify speech coherence and assess thought disorder severity, supporting automatic symptom detection in |
| | | | | | | | | | natural settings and ongoing clinical management. |
| Lee et al., (2022)[114] | United States | Sample size not detailed, targeting patients with schizophrenia and their family caregivers. | Natural Language Data (conversation content based on interactions with family caregivers) | Dialogue Management Techniques (RASA): Extract dialogue features using intent recognition and dialogue management frameworks like RASA. | RASA Framework (NLU models) | Acc: 88.31%; Prec: 89.80%; Rec: 88.22%; F1: 88.65% | Not available. | Psychosocial support or intervention, Functional training | Providing AI-based interactive training for family caregivers, informed by cognitive-behavioral therapy principles, to enhance coping strategies and psychosocial support for patients. |
| Ciampelli et al., (2023)[115] | The Netherlands | 93 patients with schizophrenia (including 40 with schizophrenia and 19 with schizoaffective disorder) and 70 healthy controls, with an average disease duration of 8.5 years, some patients in remission. | Audio Data (obtained through semi-structured interviews) | Noise Reduction and Audio Segmentation: Use PRAAT for noise reduction and segmentation, employ manual transcription (CLAN-CHILDES guidelines) and validated Kaldi ASR for automatic transcription; compute semantic similarity via Word2Vec. | RF | Acc: 76.7% (auto), 79.8% (manual); Sensitivity: 70%; Specificity: 86% | Feature Importance Analysis: Performed feature importance analysis using RF. | Symptom monitoring and assessment | Integrating automatic speech recognition and natural language processing to evaluate semantic coherence in speech, aiding in symptom assessment and supporting diagnostic processes. |
| Jeong et al., (2023)[116] | Canada | 7 hospitalized patients diagnosed with schizophrenia, | Natural Language Data (patients' verbal language records) | Language and Semantic Feature Extraction (COVFEFE, BERT): Use COVFEFE for lexical | SVM, Decision Tree, Neural Networks | AUC: ≤1.0; F1: 82% (impoverished language severity) | Feature Correlation Analysis: Analyzed the | Symptom monitoring and assessment | Applying NLP techniques to analyze linguistic markers for objective symptom |
| | | average age 37.7 years. | | richness/syntactic complexity and BERT for semantic similarity/surprisal features. | | | correlation between extracted language features (e.g., syntactic complexity, lexical richness) and clinical scores (SAPS/SANS/TLC) using Spearman correlation tests, identifying significant features through Bonferroni correction. | | evaluation in hospitalized psychiatric patients, supporting continuous and accurate symptom monitoring. |
| Chan et al., (2023)[117] | United States | 167 patients with schizophrenia and 90 healthy controls, in clinical stable phase. | Autobiographical Narrative Text Data | Sentence Embedding (USE): Embed sentences using the Universal Sentence Encoder. | SVM and LR | AUC: 0.80 (SVM), 0.78 (LR) | Feature Importance Analysis: Conducted feature importance analysis using Linear SVM and LR. | Symptom monitoring and assessment | Capturing language-based self-experience features correlated with clinical symptoms, offering opportunities for symptom tracking and progression assessment. |
| Just et al., (2023)[118] | Germany | 71 patients (51 with schizophrenia and 20 with schizoaffective disorder), with an average disease duration of 13.47 years. | Oral Corpus Data (patients' oral narratives) | Corpus Preprocessing and Word Vector Construction: Remove stopwords, perform lemmatization, and construct sentence vectors from word vectors. | NLP Models (Word2Vec, GloVe) for semantic coherence | r: Correlated with clinical scores; Word2Vec Coherence: Negative correlation with negative/disorganized symptoms | Not available. | Symptom monitoring and assessment | Using AI for semantic analysis to detect incoherence, aiding clinicians in monitoring symptom severity and changes in patients' mental states. |
| Shibata et al., (2023)[119] | Japan | 18 patients with schizophrenia (GAF score range 32–70, average age 45.17 years). | Audio Data | Voice Feature Extraction (Librosa): Extract voice features using Librosa (180-dimensional vectors). | k-NN, RF, SVR among ten ML algorithms | RMSE: Mean=13.433; best model=k-NN | Feature Importance Analysis: Assessed feature importance using SHAP values, with MFCC features contributing the most to the model. | Symptom monitoring and assessment, Psychosocial support or intervention | Estimating quality-of-life dimensions through voice feature analysis, enabling remote and continuous monitoring to support long-term schizophrenia rehabilitation management. |
| Wang et al., (2023)[120] | United States | 384 patients with schizophrenia, with auditory verbal hallucination symptoms. | Audio Data (voice diaries), Text Data (voice transcriptions), and Sensor Data (GPS, phone usage, activity monitoring) | Multimodal Feature Extraction: Combine voice (frequency, loudness), language (LIWC), and sensor data features. | Bidirectional GRU (BiGRU) and Self-Attention Mechanism | Weighted F1: ≤0.84 (audio, text, sensor data) | Feature Importance Analysis: Evaluated feature contributions using SHAP. | Symptom monitoring and assessment | Combining voice diaries and multimodal data with deep learning models to assess auditory verbal hallucination severity, facilitating targeted symptom monitoring and management. |
| Bain et al., | United States, | 431 stable | Video Data (used to | Facial Recognition and | Facial | Pharmacokinetic | Indirect | Medication | Enhancing medication |

| (2017)[121] | Russia, United Kingdom | non-smoking patients with schizophrenia. | verify medication intake) | Computer Vision: Apply facial recognition and computer vision algorithms to extract facial features. | Recognition and Computer Vision with an AI Platform (details not given) | Adherence: 89.7% (AI) vs. 71.9% (mDOT); r: 0.33 | Interpretability: Demonstrated model interpretability indirectly by monitoring video behaviors to detect abnormal medication adherence. | management | adherence by identifying non-adherent patients, providing timely interventions, reducing relapse risk, and improving overall treatment outcomes. |
|---|---|---|---|---|---|---|---|---|---|
| Shen et al., (2021)[122] | China, United Kingdom | 281 patients with chronic schizophrenia (average disease duration 34.54 years) and 35 healthy controls. | Image Data (art therapy images) | Color Distribution and Brush Stroke Analysis: Use color histograms and Hough transform to analyze color distribution and brush stroke patterns. | CNN | RMSE (PANSS): Total=15.39; Positive=4.36; Negative=7.24; General Psychopathology=6.48 | Not available. | Symptom monitoring and assessment | Using deep learning analysis of patient-generated drawings to predict PANSS scores, assisting in objective symptom severity assessment. |
| Cotes et al., (2022)[123] | United States | 50 patients with schizophrenia, 50 patients with depression, and 50 healthy control subjects. | Multimodal Data (video, audio, motion data [accelerometer], heart rate data, electroencephalogram) | Multimodal Data Analysis: Extract features from multimodal data (facial expressions, voice analysis, heart rate variability). | Deep Learning Models (CNNs) and traditional ML models (RF) | Performance Metrics: Not available | Not available. | Symptom monitoring and assessment, Psychosocial support or intervention | Employing facial expression and voice data to evaluate emotional fluctuations and symptom severity, with multimodal fusion helping predict patient subtypes and functional changes. |
| Martin et al., (2022)[124] | Germany, Canada | 20 patients with schizophrenia | Motion Capture Data (gait, full-body | Fourier Decomposition (FD) and PCA: Perform | Motion Capture | Cohen’s d: 0.6–1.5 | Not available. | Symptom monitoring | Utilizing 16 defined movement markers to |
| | | spectrum disorders and 20 healthy control subjects. | movement characteristics) | FD and PCA for dimensionality reduction. | (MoCap) and Linear Discriminant Function (LDF) | | | and assessment | monitor motor function changes in schizophrenia, informing individualized rehabilitation strategies and functional recovery plans. |
| Koppe et al., (2019)[125] | Germany, The Netherlands, United States | Patients with schizophrenia, sample size not explicitly reported. | Behavioral Data, Physiological Data, Environmental Data (GPS tracking, sensor data), and Psychological Scores | Multimodal Data Integration: Integrate multiple data types for feature extraction. | RNN | Performance Metrics: Not available | Not available. | Risk management, Symptom monitoring and assessment, Psychosocial support or intervention | Using recurrent neural network analyses of daily behavioral and physiological data to predict symptom trajectories and identify optimal intervention windows. |
| Umbricht et al., 2020[126] | United States | 33 patients with chronic schizophrenia, stably taking antipsychotic medication. | Wearable Device Data (activity data collected via wristbands) | Activity and Gesture Feature Extraction: Use accelerometer signals to derive activity and gesture features. | 9-Layer Convolutional Recurrent Neural Network | Activity Recognition Acc: >95% | Not available. | Symptom monitoring and assessment | Inferring negative symptom severity (e.g., avolition, expressive deficits) from activity data, aiding in continuous symptom assessment. |
| Adler et al., (2020)[127] | United States | 60 patients with schizophrenia, schizoaffective disorder, and | Passive Sensing Data (acceleration, phone app usage, calls, location | Neural Network Feature Extraction: Utilize encoder-decoder architectures | Encoder-Decoder Neural Networks (Fully Connected | Median Sensitivity: 0.25 (IQR 0.15–1.00); Specificity: 0.88 (IQR 0.14–0.96); NR30 ↑108% | Feature Importance Analysis: Quantified the | Risk management | Predicting relapse risk through passive sensing data, assisting clinicians in |
| | | non-specified psychosis (42 without relapse and 18 with relapse), including patients with multiple relapses. | information, screen activity, sleep patterns, etc.) | (autoencoders, GRU sequence models) for feature extraction. | Autoencoders, GRU-based) | (30 days pre-relapse) | discriminative ability of features under different conditions (e.g., approaching relapse period vs. healthy period) using effect sizes (Cohen's d) to determine feature importance. | | recognizing early warning signs and implementing timely interventions. |
| Nguyen et al., (2022)[128] | Taiwan, Vietnam | Two public datasets (Psykose and Depresjon), including 22 patients with schizophrenia/23 patients with mood disorders and 32 healthy control subjects. | Wearable Device Activity Signals | Time-Series Segmentation and Transformation: Convert raw time-series data into "pseudo-images" for feature extraction. | CNNs (VGG16, ResNet50v2, XceptionNet, EfficientNetB1); RNNs (LSTM, GRU, Attention-based LSTM/GRU) | ResNet50v2 (Binary): Acc: 0.88; Sensitivity: 0.93; F1: 0.88; XceptionNet: Acc: 0.86; Sensitivity: 0.92; F1: 0.87 | Not available. | Symptom monitoring and assessment | Applying wearable device activity signals and deep learning models for early diagnostic support, enhancing rehabilitation management and symptom monitoring. |
| Zlatintsi et al., (2022)[129] | Greece | 39 patients with mental disorders, including 12 with schizophrenia. | Smartwatch Sensor Data (physiological data such as heart rate variability; behavioral data | Physiological and Behavioral Feature Extraction (Signal Processing): Apply signal processing to extract | RF for multimodal analysis; Transformer, GRU, CNN, and | Acc, Sensitivity, Specificity: Good relapse prediction (no exact values) | Feature Importance Analysis: Assessed feature | Risk management, Symptom monitoring | Combining smartwatch-derived physiological data and patient interview videos with machine |
| | | | such as step counts, exercise intensity, gait changes), Video Data (facial expressions), and Natural Language Data (interview recordings) | physiological/behavioral features, analyze facial expressions with deep learning, and measure pauses, speech rate, pitch changes; perform feature fusion. | Fully Connected Network (FNN) for relapse detection; C3D and Two-Stream Network for video analysis; Autoencoder for anomaly detection | | importance based on the attention mechanisms within the Transformer architecture and quantified feature importance through univariate perturbations. | and assessment | learning methods to predict relapse risk and track psychological symptom changes. |
| Lin et al., (2023)[130] | United States | Alex Street psychotherapy dataset, containing over 950 transcribed psychotherapy sessions, with participants including multiple anonymous therapists and patients (including those with schizophrenia). | Natural Language Data (transcriptions of psychotherapy dialogues between therapists and patients) | Real-Time Voice Segmentation, Text Embedding, Topic Modeling, and Dimensionality Reduction: Conduct speaker diarization, apply Doc2Vec embeddings, use embedded topic modeling, and employ PCA and UMAP for dimensionality reduction. | RL (DDPG, TD3, BCQ), Recommendation Algorithms (Content-Based, Collaborative Filtering) | WAI Scores: DISMOP-BCQ-GOAL=0.6424 (best) | Visual Explanation: Visualized model behavior using PCA and transition matrices. | Psychosocial support or intervention | Exploring AI-based psychotherapy recommendation systems to enhance therapist-patient interactions, supporting symptom alleviation, patient management, and improvements in social functioning. |

*Note:* The following abbreviations are used throughout this paper: Support Vector Machine (SVM); Random Forest (RF); Logistic Regression (LR); Gradient Boosted Regression Trees (GBRT); Gradient Boosting Machine (GBM); k-Nearest Neighbors (k-NN); Convolutional Neural Network (CNN); Recurrent Neural Networks (RNN); Gated Recurrent Unit (GRU); Reinforcement Learning (RL); Accuracy (Acc); Area Under the

Curve (AUC); Receiver Operating Characteristic (ROC); Mean Absolute Error (MAE); Pearson Correlation Coefficient (r); Root Mean Square Error (RMSE); Mean Squared Error (MSE); Mean Relative Error (MRE); Sensitivity (also TPR, True Positive Rate); Specificity (also TNR, True Negative Rate); F1 Score (F1); Precision (Prec); Recall (Rec); Confidence Interval (CI); Interquartile Range (IQR); Principal Component Analysis (PCA); Natural Language Processing (NLP); Linguistic Inquiry and Word Count (LIWC, a lexicon-based language analysis tool); Term Frequency–Inverse Document Frequency (TF-IDF); Recursive Feature Elimination (RFE); Gaussian Mixture Model (GMM); Dynamic Time Warping (DTW); Word Mover's Distance (WMD); Automatic Speech Recognition (ASR); Bidirectional Encoder Representations from Transformers (BERT); Automated Machine Learning (AutoML); Fourier Decomposition (FD); Uniform Manifold Approximation and Projection (UMAP); SHapley Additive exPlanations (SHAP, a method for feature importance evaluation); Generalized Estimating Equations (GEE, a statistical method for correlational analysis); Mel-frequency Cepstral Coefficients (MFCC, commonly used in voice and speech analysis).

Results

*Publication trends and geographical distribution.* Figure 1a illustrates the publication trends and geographical distributions of the studies included in this review. A total of 61 studies were identified between 2012 and January 2024. Since 2020, the number of studies has rapidly increased, with studies published in 2022, 2023, and up to January 2024 accounting for 81.97% of the research.

In terms of geographical distribution, most studies were conducted in the United States (n=31; 50.82%), followed by the United Kingdom (n=7; 11.48%). Beyond these, 18 additional countries contributed to this area of research, including (but not limited to) China, Singapore, Greece, South Korea, the Netherlands, and Canada (Table 1).

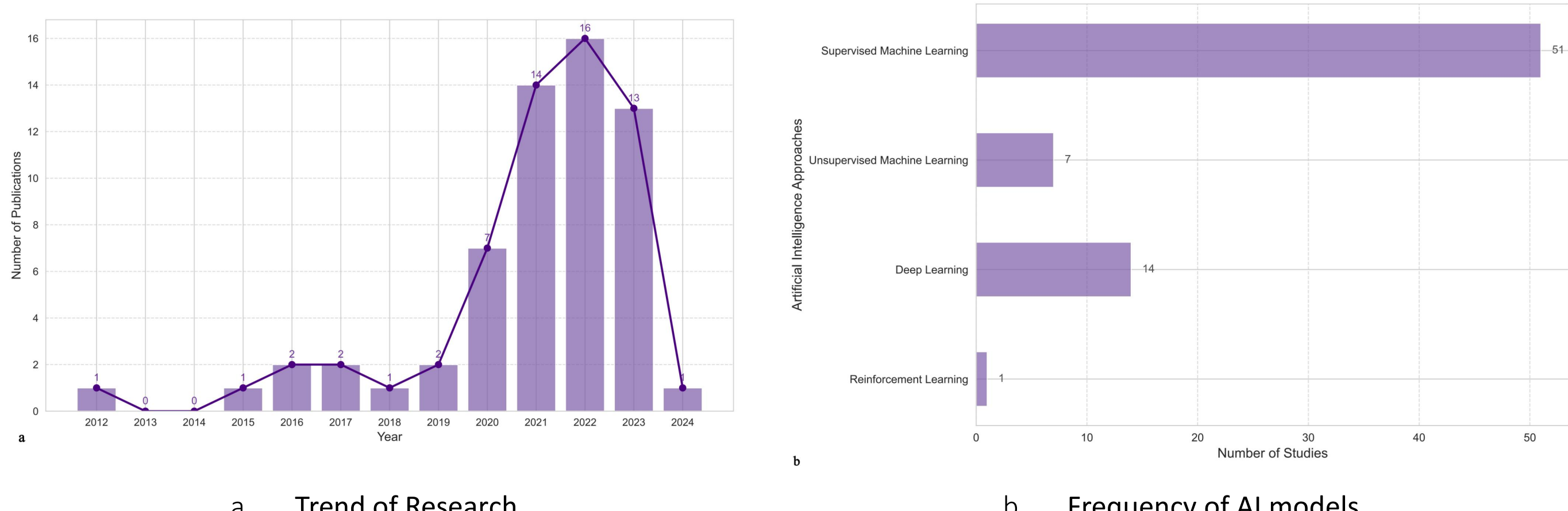

a. Trend of Research b. Frequency of AI models

Figure 1 Trends and Applications of AI in Studies.

*Sample characteristics.* The studies included encompassed over 49,000 participants diagnosed with schizophrenia, with 49 studies explicitly reporting the number of patients with schizophrenia in their datasets (Table 1). The distribution of sample sizes across these studies was categorized as follows: 23 studies included small sample sizes (≤50 participants), 10 featured medium sample sizes (51-200 participants), and 16 included large sample sizes (>200 participants).

Additionally, some studies have incorporated established public datasets that include patients with schizophrenia, such as the CrossCheck [100], Psykose [110, 128], and Depresjon [128], as well as the Alex Street psychotherapy dataset with a substantial corpus of over 950 transcribed sessions [130].

*Data types.* The included studies employed a diverse range of data types to develop AI models for schizophrenia rehabilitation management (Table 1). A total of 23 studies utilized textual and language-based data, encompassing conversation transcripts [70, 114, 117, 118, 129, 130], speech recordings or audio-based linguistic features [78, 82, 87, 103, 112, 115, 116, 119, 120, 123, 129], social media postings [74], internet search queries [81], transcribed psychotherapy sessions [99, 104, 103], and various narrative texts or interview-based transcriptions [88, 91, 103, 113, 114, 116, 117, 118, 129, 130].

Twenty studies incorporated sensor and physiological data including heart rate (ECG) measurements [71, 85, 92, 123, 129], accelerometer-derived physical activity metrics [71, 84, 85, 92, 97, 108, 110, 111, 120, 123, 136-139], and GPS-based location tracking [72, 83, 84, 92, 97, 108, 111, 120, 125, 127, 129].

Thirteen studies included behavioral data reflecting user interactions such as smartphone usage patterns [72, 73, 83- 85, 92, 99, 100, 120, 123, 127, 129], call and SMS logs [73,84,92], and activity frequency [84, 100, 120]. Environmental data, such as ambient light intensity and sound levels, were identified in five studies [72, 73, 83, 100, 111]. Self-reported data, including Ecological Momentary Assessment (EMA) questionnaires and Likert scale evaluations, were present in five studies [73, 76, 77, 97, 115].

Multimedia data were employed in nine studies, encompassing video-based assessments (eye tracking, facial expressions) [90, 91, 103, 123, 129] and image-based evaluations (facial emotion recognition tasks, art therapy images) [95, 96, 122, 123]. Finally, EHR and related clinical record data were utilized in five studies [75, 86, 101, 102, 107], providing diagnostic records, prescription information, hospitalization data, and other longitudinal clinical parameters.

*Feature engineering and representation learning.* The included studies employed various feature engineering and representation learning techniques to preprocess and transform the data for AI models (Table 1). Data preprocessing methods were detailed in five studies: aggregation, normalization (zero-mean and unit-variance), one-hot encoding, min-max scaling, and imputing missing data using k-nearest neighbors (k-NN) [72, 86, 99, 101, 102]. Dimensionality reduction and feature selection techniques were employed in 14 studies to enhance model efficiency and identify the most relevant features for training. These methods included PCA, AutoEncoders (AE), and UMAP for dimensionality reduction [84, 86, 105, 109, 124, 130], as well as CfsSubsetEval, M5 Prime, Pearson correlation, ElasticNet, GINI index, Boruta, decision tree-based feature evaluation, and AutoML-based methods for feature selection [70, 73, 89, 91,93-95, 97, 101, 107, 110].

The physiological signals and textual data were extensively processed to extract meaningful features and generate structured representations of the AI models. Signal processing techniques were used in seven studies to extract features from speech and physiological signals, such as pitch, jitter, shimmer, and heart rate variability (HRV), enabling the analysis of behavioral and physiological patterns [71, 78, 82, 83, 87, 119, 129]. Similarly, textual and language representation learning was applied in 12 studies utilizing methods such as n-gram models, LIWC, Word2Vec, BERT, TF-IDF, and Doc2Vec to generate vector representations of textual data. These representations supported downstream tasks, including sentiment analysis, symptom tracking, and interactive interventions [70, 74, 81, 88, 105, 112, 115, 116, 117, 118, 120, 130].

Multimodal fusion techniques were applied in four studies to integrate diverse data sources including facial expressions, voice features, and sensor-based behavioral data. These techniques enable a more comprehensive representation of patient behavior and emotional states, allowing for a more holistic analysis of rehabilitation progress and needs [120, 123, 125, 129].

*Model.* As shown in Figure 1b, the included studies primarily employed supervised ML methods (n = 51; 83.6%). Unsupervised approaches (n = 7) were also reported, notably involving clustering methods such as k-means, fuzzy clustering, and Gaussian Mixture Models (GMM). DL methods (n = 14), including CNNs, RNNs, and AE, feature prominently in various tasks such as feature extraction, classification, and pattern recognition. RL methods (n = 1) were identified in a limited number of studies (Table 1).

Supervised machine-learning models were most frequently utilized in the included studies. RF (n = 21) [72, 74, 80, 81, 84, 86-89, 93, 95, 97, 98, 101, 103, 110, 111, 115, 119, 123, 129], and SVM (n = 14) [70, 71, 76, 80-83, 88, 89, 94, 95, 116, 117, 119] were the predominant algorithms applied. Gradient Boosting methods (GBRT/GBM) (n = 4) [73, 81, 100, 101] and Logistic Regression (n = 2) [80, 86] were also commonly employed. XGBoost (n = 3) [84, 88, 102] and Decision Trees (DT) (n = 4) [70, 75, 91, 116] were used less frequently. Additionally, Lasso Regression (n = 3) [78, 97, 101] and multioutput Support Vector Regression (m-SVR) (n = 1) [83] were utilized in specific contexts to enhance model performance.

Supervised ML tasks predominantly focused on classification rather than regression. Classification models were used extensively in the reviewed studies. By contrast, regression-based approaches were comparatively rare, identified in association with Lasso Regression (n = 3) [78, 97, 101], Linear Regression [80, 86], and m-SVR (n = 1) [83].

Unsupervised learning techniques, including clustering methods, such as k-means (n = 2) [105, 108], fuzzy clustering (n = 1) [109], and GMM (n = 1) [111], as well as other structural learning or representation methods (n = 3, e.g., HMM [77], GFCI [106], and word embedding [118]), were employed to uncover intrinsic patterns within the data.

The DL method was used in 14 studies. These include CNNs [122, 123, 128, 129], RNNs (including variants such as GRU and BiGRU) [120, 125-129], transformers (e.g., BERT) [113-116, 129], and AE [86, 127]. In seven studies [86, 88, 95, 113, 116, 123, 129], ML and DL models were employed, highlighting their complementary strengths in data analysis. The primary strategies involved using ML models (e.g., RF and SVM) to process structured data or perform initial classification while leveraging DL models (e.g., CNNs, Transformer) to extract complex nonlinear features. Alternatively, DL models were utilized to learn high-dimensional feature representations of samples, followed by ML models for classification or regression tasks. RL methods were identified in one study [130] utilizing algorithms such as DDPG, TD3, and BCQ.

*Model interpretability.* The included studies employed various techniques to enhance the interpretability of AI models for schizophrenia rehabilitation (Table 1). Interpretability is critical for understanding how models make decisions and for identifying relationships between target variables (e.g., symptom severity and relapse risk) and input features (e.g., behavioral metrics and physiological signals). The clarity provided by interpretability is essential to ensure model reliability and establish trust among clinicians, patients, and stakeholders in the mental health context.

Two primary strategies for feature interpretability were identified: intrinsic interpretability and post hoc explanations. Intrinsic interpretability utilizes the properties of tree-based models, such as RF and DT, to rank input variables by their influence on model decisions. This intrinsic interpretability approach was the most commonly applied approach used in 15 studies to identify key factors associated with rehabilitation-related tasks, including symptom monitoring, treatment adherence, and relapse risk prediction [72-75, 79, 81, 87, 91, 93, 95, 97, 101, 103, 107, 115].

Post hoc explanations, such as SHAP, were employed in five studies [98, 99, 102, 119, 120]. In addition to ranking feature importance, SHAP values provide deeper insights by quantifying the direction and magnitude of each feature's contribution to individual predictions. This capability allows researchers to interpret complex non-linear models and better understand nuanced feature interactions, thereby enhancing model transparency and applicability in schizophrenia rehabilitation. Beyond SHAP, other post hoc methods included visual explanations (e.g., class contrast analysis and heatmap generation) in three studies [86, 92], causal path explanations in one study [106], and attention-based interpretability in one study [129].

Additionally, five studies employed correlation or statistical analyses to elucidate the relationships between features and clinical outcomes [82, 84, 116, 121, 127], whereas one study leveraged indirect interpretability by monitoring behavioral patterns in video data [121].

*Rehabilitation application areas.* AI models in schizophrenia rehabilitation predominantly relied on multimodal data inputs, including speech patterns, behavioral metrics, physiological signals, and environmental contexts (Table 1). These data were processed through various ML, DL, and, in some cases, reinforcement learning frameworks. AI-driven systems often operate in real-time, enabling early warnings, personalized interventions, and dynamic treatment optimization to improve patient outcomes and quality of life.

Symptom Monitoring and Assessment emerged as the most common application area, covered by 47 studies [71-74, 76-78, 81-85, 87-92, 94-100, 102, 104, 105, 108-113, 115-126, 128, 129]. AI models in this domain integrated continuous patient data, such as speech transcripts, wearable sensor data, and smartphone logs, to assess clinical states. These models aim to correlate real-time behavioral data with clinical measures, such as symptom scales. For example, Wang et al. [73] combined GBRT models with passive sensing data (e.g., activity levels and environmental cues) to predict weekly scores on a Brief Psychiatric Rating Scale (BPRS). The model achieved a Pearson's r of 0.70, allowing for early detection of symptom escalation and timely intervention. Similarly, Holmlund et al. [112] adopted a language-centric approach using linear regression and automatic speech recognition (ASR) to correlate daily speech-based memory test performance with cognitive states. This language-centric model achieved a near-human-level correlation (r = 0.83), enabling tracking of subtle cognitive shifts over time, providing treatment adjustments, and potentially improving long-term functional outcomes.

Risk Management was addressed in 16 studies [72, 75, 79-81, 83, 85, 86, 92, 93, 107, 109, 111, 125, 127, 129] focusing on predicting relapse risk and violence risk. These studies typically employed binary classification models, incorporating positive and negative samples (e.g., relapse events and violent incidents) to train the models effectively. Wang et al. [80] used clinical, demographic, and sociocultural data from 275 patients with schizophrenia to predict violent behaviors. Random Forest achieved the best performance, with an accuracy of 62% and an AUROC of 0.63 in predicting violent tendencies.

Medication Management was the focus of eight studies [70, 79, 86, 101, 102, 104, 107, 121], which primarily utilized supervised machine learning models to assess and predict medication adherence. These strategies primarily involved patient speech analysis, adherence logs, and prescription histories to predict medication adherence and therapeutic effectiveness. Howes et al. [70] used SVM and DT on conversational data, extracting discourse features (e.g., turn-taking frequency and question counts) and symptom- or treatment-related keywords to predict medication adherence, achieving accuracies of 91.1%–93.0%. Jeon et al. [101] leveraged a large-scale national insurance database with over 5,000 cases of schizophrenia by applying RF and GBM to integrate prescription histories, diagnostic records, and care utilization patterns. Their results showed that patients receiving AI-recommended treatment strategies maintained medication continuity at rates 1.2–1.5 times higher than those without AI guidance.

Functional Training was examined in five studie[s] [89, 90, 91, 106, 114] focusing on improving patients' cognitive and social skills through adaptive intervention. In this context, AI tools were employed to predict or enhance outcomes in cognitive and social training tasks, enabling dynamic adjustments to intervention strategies based on user performance and environmental cues. For example, Lee et al. [114] applied the RASA NLU to assist family caregivers by simulating interactive dialogues grounded in cognitive–behavioral therapy principles. This application of the RASA provided real-time adaptive guidance that helped caregivers refine their communication approaches, thereby indirectly improving patient rehabilitation experiences within home settings.

Psychosocial Support or Intervention was addressed in seven studies [84, 99, 114, 119, 123, 125, 130] focusing on delivering timely feedback and interventions aimed at enhancing patient engagement, emotional well-being, and therapeutic outcomes. AI-driven approaches include conversational agents, topic recommendation mechanisms, and adaptive dialogue models to predict or improve the effectiveness and participation rates of psychological interventions. Lin et al. [130] integrated DDPG, TD3, and BCQ to develop a psychotherapy topic recommendation system that facilitates meaningful therapist–patient interactions.

## Discussion

This systematic review included 62 studies published between 2012 and January 2024, with a significant increase in research activity observed after 2020, and a predominant focus on studies conducted in the United States. Most of these studies employed supervised ML techniques, with RF, SVM, GBRT, and XGBoost as the most frequently applied models. These models were often integrated with multimodal data sources including speech, behavioral metrics, sensor signals, and environmental measures. To enhance transparency and interpretability, many studies utilized feature importance analysis methods, and techniques such as SHAP value analysis and RF-based feature importance ranking being widely applied.

Symptom monitoring and assessment (47 studies) and risk management (16 studies) have emerged as the most extensively explored domains, reflecting the critical role of AI in predicting relapse risk and dynamically tracking clinical states. Medication management (eight studies) and psychosocial support or intervention (seven studies) were also widely investigated, whereas functional training received comparatively less attention — likely due to the complexity of standardizing real-world tasks, challenges in data capture for interactive interventions, and the nascent stage of AI-based functional training solutions. These findings underscore the progress and diverse applications of AI in schizophrenia rehabilitation, demonstrating its potential to optimize interventions, improve patient outcomes, and support personalized care. However, this review discusses the limitations and promising research directions across four dimensions: data, models, application areas, and application practices.

Regarding data-related aspects, limitations in data quantity and quality remain primary challenges in applying AI to schizophrenia rehabilitation. Among the included studies, 67.3% reported datasets with < 200 participants, underscoring the prevalence of small sample sizes. In addition to well-known problems such as small sample sizes and uneven data distributions [131-132], there remains a scarcity of clinically significant negative samples such as instances of relapse or non-adherence to treatment plans [133-134]. These limitations hinder the ability to draw broad conclusions and accurately predict rare outcomes that are critical for tasks, such as relapse risk prediction, medication adherence prediction, and violence risk assessment. Some studies combined rehabilitation phase data with diagnostic records, demonstrating that multisource data integration can enrich contextual information and enhance model robustness [93]. This multisource data integration strategy appears promising for improving predictive accuracy and warrants broader implementation. Beyond conventional data modalities, the use of social media data (e.g., Reddit and Twitter) is a rapidly evolving area [135-137]. Some studies included in our review have already explored these sources, particularly for symptom monitoring and risk prediction [138-139]. This approach of utilizing social media data can capture real-time behavioral and emotional indicators, thereby enhancing the responsiveness and personalization of rehabilitation interventions.

With regard to modeling, supervised machine learning was widely adopted, particularly for its efficiency in handling structured data and its ability to be integrated with interpretable techniques such as feature importance analysis (e.g., SHAP and RF rankings). These methods enable clinicians to identify key variables that are most closely linked to symptom severity, relapse risk, and medication adherence patterns. By focusing on these critical factors, interventions can be directed more effectively toward areas of greatest clinical relevance. However, prevalent reliance on pre-extracted features, particularly in textual or physiological datasets, can obscure valuable contextual signals [140-141]. For example, in language data, subtle shifts in a patient's discourse structure or emotional tone may indicate emerging psychotic symptoms [142], whereas in physiological signals, sustained patterns across time and environmental conditions can reveal stressors that contribute to relapse. Loss of this contextual richness limits the model's capacity to fully capture the complexity of patient experiences.

Deep learning offers significant advantages over traditional feature engineering combined with machine-learning approaches, particularly when handling text or physiological signals. Its ability to process high-dimensional, multimodal data directly—without extensive manual preprocessing—enables models to capture subtle temporal, linguistic, and environmental patterns [143-144]. This capability has been consistently demonstrated in previous studies, in which deep learning models outperformed traditional methods in providing accurate and context-aware predictions [145-146]. Architectures including CNNs, RNNs, Transformers, and tabular data-focused models, such as TabNet [147], can fuse speech, visual signals, and sensor data to uncover latent patterns in patient trajectories. Although the "black-box" nature of deep learning models remains a concern [148-149], recent advances in explainability techniques, such as saliency maps, attention weight visualization, and layer-wise relevance propagation [150-152], offer promising solutions to enhance model transparency and interpretability.

Recent advances in LLMs have demonstrated significant potential for a range of healthcare applications, particularly in mental health [153]. Tasks such as symptom analysis [154], diagnostic support [155-156], and educational therapeutic intervention [157] utilize LLMs. These capabilities closely align with the key domains identified in this review, including risk management and symptom monitoring. Additionally, conversational agents powered by LLMs (e.g., OpenAI's Advanced Voice Mode) [158] can dynamically tailor psychoeducational content, suggest targeted cognitive exercises, and refine communication strategies. Regarding the final cutoff date for our literature review, we did not identify any studies that applied LLMs directly to rehabilitation scenarios for patients with schizophrenia. However, considering the rapid advancements in LLMs and the increasing application of such models in mental health tasks, we believe that LLMs can potentially contribute meaningfully to the rehabilitation of patients with schizophrenia, particularly in specific areas.

However, a major problem with LLMs outputs is hallucinations, in which models may generate unpredictable, inaccurate, or clinically irrelevant content. These risks underscore the need for rigorous validation and robust safeguards to ensure safe application in sensitive contexts[159-160]. These concerns are increasingly recognized at the regulatory level, for example, in the EU AI Act[161], which emphasizes the need for transparency and adherence to safety standards. Despite these challenges, the substantial potential of LLMs in enhancing patient interactions and providing nuanced therapeutic support is widely acknowledged[155, 162-163]. Future research should prioritize establishing rigorous domain-specific evaluation benchmarks that capture the distinct complexities of mental health contexts. Complementing these benchmarks with targeted fine-tuning approaches and integrating LLMs into rehabilitation-specific frameworks can enhance their reliability and clinical utility, particularly for supporting personalized care and therapeutic interventions in schizophrenia rehabilitation.

Most current interventions for schizophrenia rehabilitation focus on passive data collection and monitoring. This focus is particularly evident in areas, such as treatment resistance, relapse risk prediction, and medication adherence. However, research on AI's role in dynamic, goal-oriented interventions remain limited. Expanding AI applications in this field could yield significant benefits, particularly in functional training and psychological health interventions. For instance, integrating AI with serious games, which are games designed for therapeutic purposes [164], allows for real-time monitoring and assessment of patients' rehabilitation status, enabling AI systems to continuously evaluate and track performance metrics. Such integration of AI with serious games facilitates the dynamic adaptation of game-based training programs to better align with individual progress and needs, as observed in one study [108] included in this review. Additionally, incorporating AI-powered virtual agents into VR-based social skills training creates interactive and adaptive simulations of real-world social situations.

Intervention practices have demonstrated substantial potential for leveraging AI to enhance schizophrenia rehabilitation, particularly in risk management and symptom monitoring. However, we also observed that the majority of the included studies originated from high-resource settings, such as the United States (n=31), the United Kingdom (n=7), Canada (n=4), and Germany (n=3). The concentration of studies in these high-resource settings reflects disparities in AI development and implementation driven by organizational, policy, and technical barriers that limit access to advanced technologies in low-resource or underserved communities[165-166]. Expanding AI-driven interventions to these communities is an important and valuable research direction, as it could help mitigate existing gaps in mental healthcare access. Key strategies for achieving this include utilizing mobile-based platforms to deliver low-cost, scalable AI solutions, such as symptom monitoring apps or telehealth systems powered by conversational agents, while addressing the lack of necessary technical infrastructure, such as reliable internet access and smart devices, in low-income and remote areas. Additionally, integrating AI into community health efforts, such as employing LLM-driven educational tools or training modules for local caregivers, could broaden the reach and impact of rehabilitation services. Policy-level support for infrastructure development and resource allocation is critical for facilitating the adoption of AI technologies in underserved settings[166-167].

This review, similar to all systematic syntheses, has some limitations. We may have overlooked some studies due to non-standard terminology or limited dissemination, despite our comprehensive search across multiple databases. Furthermore, the heterogeneity in AI architectures, data modalities, clinical populations, and outcome measures across the included studies could have constrained our ability to perform direct comparisons or conduct robust effect-size syntheses, particularly given the rapidly evolving nature of AI-driven interventions. While our focus on English-language publications was primarily driven by resource constraints, this choice may limit the global perspective on AI-based interventions. Future reviews should incorporate a multilingual literature to achieve a more globally representative perspective on AI applications in schizophrenia rehabilitation.

In summary, AI holds substantial promise for advancing schizophrenia rehabilitation by providing tools for personalized symptom monitoring, relapse prevention, medication management, and psychosocial support. However, problems persist in data, modeling, and connecting with community-based services. Nonetheless, new developments in deep learning architectures, LLM-driven agents, and interactive digital interventions have shown that things can improve significantly in the future. By fixing these problems, encouraging collaboration across disciplines, and improving data sharing, usability, and the design of interventions, AI-driven solutions can increasingly improve traditional rehabilitation methods, ultimately leading to better patient outcomes and quality of life.

## Methods

When discussing the application of AI techniques in schizophrenia rehabilitation management, complex models, data, and application scenarios were encountered. Given the significant differences in objectives, technology, and evaluation metrics among these studies, this study employed a systematic scoping review methodology. This method allows us to comprehensively review and synthesize existing research findings rather than compare quantitative results from homogeneous studies. This methodological strategy was designed to broadly evaluate the application of AI in schizophrenia rehabilitation and delve into its potential value and challenges. This study strictly adhered to the Preferred Reporting Items for Systematic reviews and Meta-Analyses extension for Scoping Reviews (PRISMA-ScR) [168].

### Search strategy

#### *Search sources*

The search strategy for this review aimed to comprehensively cover research on AI applications in schizophrenia rehabilitation, ensuring the breadth and depth of the studies. In January 2025, we searched four electronic databases: PubMed, Web of Science, IEEE Xplore, and ACM Digital Library. PubMed is a comprehensive medical database that covers most mental rehabilitation studies. The top engineering and computer science databases, Web of Science, provide interdisciplinary research literature, IEEE Xplore, and the ACM Digital Library, and contain a wealth of technical and developmental literature on AI models and their applications. Additionally, we reviewed the titles and abstracts of the first ten pages of Google Scholar results (approximately the first 100 results) to identify potentially relevant studies. Only those studies that appeared to be the most relevant to the topic were subsequently included in the full review, ensuring a comprehensive yet focused selection of literature.

Studies published between 2012 and the present review were included. This time frame was selected on the basis of several considerations. In 2012, AlexNet introduced an epoch-making moment in AI, opening up a new era in DL research and applications [169]. Since then, AI has undergone rapid revolutionary development, particularly in natural language processing (NLP) and computer vision [170]. Second, according to previous and current literature reviews, research on AI techniques for mental disorders before 2012 was relatively scarce. We performed backward and forward reference checks to ensure the comprehensiveness and depth of the review. This process included screening the reference lists of the included publications (backward checking) and identifying the subsequent studies that cited them (forward checking) to identify potentially missing relevant studies.

#### *Search terms*

We meticulously designed the search terms under the guidance of two experts in mental health rehabilitation to accurately identify relevant research. The search terms included target population, target technologies, and research scenarios. The keyword search strategy was as follows: (“artificial intelligence” OR “AI” OR “machine learning” OR “deep learning” OR “neural networks” OR “natural language processing” OR “computer vision” OR “computational intelligence” OR “data mining” OR “predictive modeling” OR “reinforcement learning”) AND (“schizophrenia” OR “schizophrenic” OR “schizoaffective disorder” OR “psychosis” OR “psychotic disorder” OR “severe mental illness”) AND (“rehabilitation” OR “recovery” OR “management” OR “care” OR “medication adherence” OR “drug compliance” OR “pharmacological management” OR “medication tracking” OR “medication optimization” OR “relapse prevention” OR “risk assessment” OR “risk prediction” OR “violence prediction” OR “crisis management” OR “cognitive training” OR “social skills training” OR “life skills development” OR “functional recovery” OR “skill-building interventions” OR “symptom tracking” OR “symptom monitoring” OR “behavioral monitoring” OR “therapeutic intervention” OR “emotional support” OR “therapy engagement” OR “psychological well-being”).

### Study eligibility criteria

This review focused on evaluating the current applications and potential of AI techniques in supporting the rehabilitation of patients with schizophrenia. The inclusion criteria required studies to involve participants with a confirmed schizophrenia diagnosis who were in a stable rehabilitation or recovery phase and to utilize intervention strategies that were suitable for home or community rehabilitation settings. In this context, rehabilitation management has been conceptualized as encompassing various functional domains essential for monitoring and managing symptoms, followed by the enhancement of cognitive and social functioning, and the eventual restoration of independent living skills and community integration.

On the basis of prior research on schizophrenia rehabilitation management [7-10] and digital interventions [171-172] and consultations with two rehabilitation specialists with over ten years of experience in schizophrenia care, we identified five rehabilitation application areas that must be addressed by the included studies:

a. Medication management (e.g., predicting medication adherence or pharmacological treatment outcomes).
b. Risk management (e.g., forecasting relapse, treatment resistance, and risk-related behaviors such as violence, self-harm, or crisis events) [173-174].
c. Functional training (e.g., cognitive and social skills training that supports community integration) [8, 175].
d. Symptom monitoring and assessment (e.g., evaluating rehabilitation progress and anticipating fluctuations in clinical status) [176].
e. Psychosocial support or intervention (e.g., providing psychological counseling, behavioral guidance, or conversation agents) [177].

The inclusion criteria required that studies involving participants with a confirmed diagnosis of schizophrenia were included. No restrictions were imposed on the basis of the research setting, comparator, country, or region of publication. Studies were excluded if:

a. Did not include data from individuals with a confirmed schizophrenia diagnosis were defined by clinical criteria, such as ICD-10, DSM-5, or equivalent diagnostic frameworks.
b. Did not involve the application of AI techniques, defined as the use of ML, DL, or RL.
c. Focused solely on non-rehabilitation application areas of AI-related schizophrenia research, such as clinical diagnostics, pathophysiological studies (e.g., neuroimaging biomarkers, genetic profiling), computational modeling for theoretical neuroscience (e.g., neural circuit simulations or cognitive model validation), drug discovery and development, disease trajectory modeling, epidemiological studies, or basic cognitive process analysis, without addressing rehabilitation outcomes or interventions.
d. Examined specialized patient subgroups differ significantly from typical community-based rehabilitation populations (e.g., forensic populations, individuals with severe criminal offenses, such as sex crimes, or those with specific comorbidities, such as advanced metabolic disorders, infectious diseases, such as COVID-19, or rare neurological conditions).
e. Relied exclusively on data types not feasible for continuous community-based rehabilitation monitoring (e.g., research-grade EEG requiring controlled environments, advanced neuroimaging techniques, genetic sequencing, or invasive neurophysiological recordings not practical for everyday settings).
f. Editorials, reviews, proposals, posters, conference abstracts, non-original research literature, and non-English publications.

Due to resource and language constraints, we excluded nine non-English studies (primarily in French and Spanish) identified via Google Scholar citations. These works primarily offered preliminary explorations or conceptual proposals without substantive empirical data on rehabilitation management. Thus, their omission likely did not materially affect the overall conclusions of our synthesis. Additionally, studies incorporating healthy control groups were eligible if their inclusion provided comparative insights into the development of AI-driven relapse detection and prevention strategies in the context of rehabilitation. These inclusion decisions were informed by the discussions and conclusions presented in the literature and feedback from mental health rehabilitation professionals, ensuring alignment with the relevant rehabilitation priorities and potential areas for AI-driven innovation.

### Study selection

The Zotero reference management tool automatically filters and removes duplicates from initial search results. Two independent reviewers (the first and second authors) conducted the initial screening on the basis of the titles and abstracts of the papers.

Subsequently, the reviewers performed a detailed full-text review of potentially eligible literature identified during the initial screening. Any disagreements during the review process were resolved through discussions between the reviewers and, if necessary, a third expert was consulted. Special attention was paid to studies whose application scenarios (i.e., whether they were used for patient rehabilitation) were not explicitly stated, and the reviewers jointly assessed potential application scenarios on the basis of the data and products. In cases of disagreement, consultations with experts in AI and mental health rehabilitation were sought to make the final decisions.

On the basis of Cohen's kappa statistic, the consistency among the reviewers was calculated to be 0.92. A total of 247 potentially relevant articles were identified. After applying the inclusion and exclusion criteria, 186 studies were excluded, leaving 61 articles that met the inclusion criteria (Figure 2).

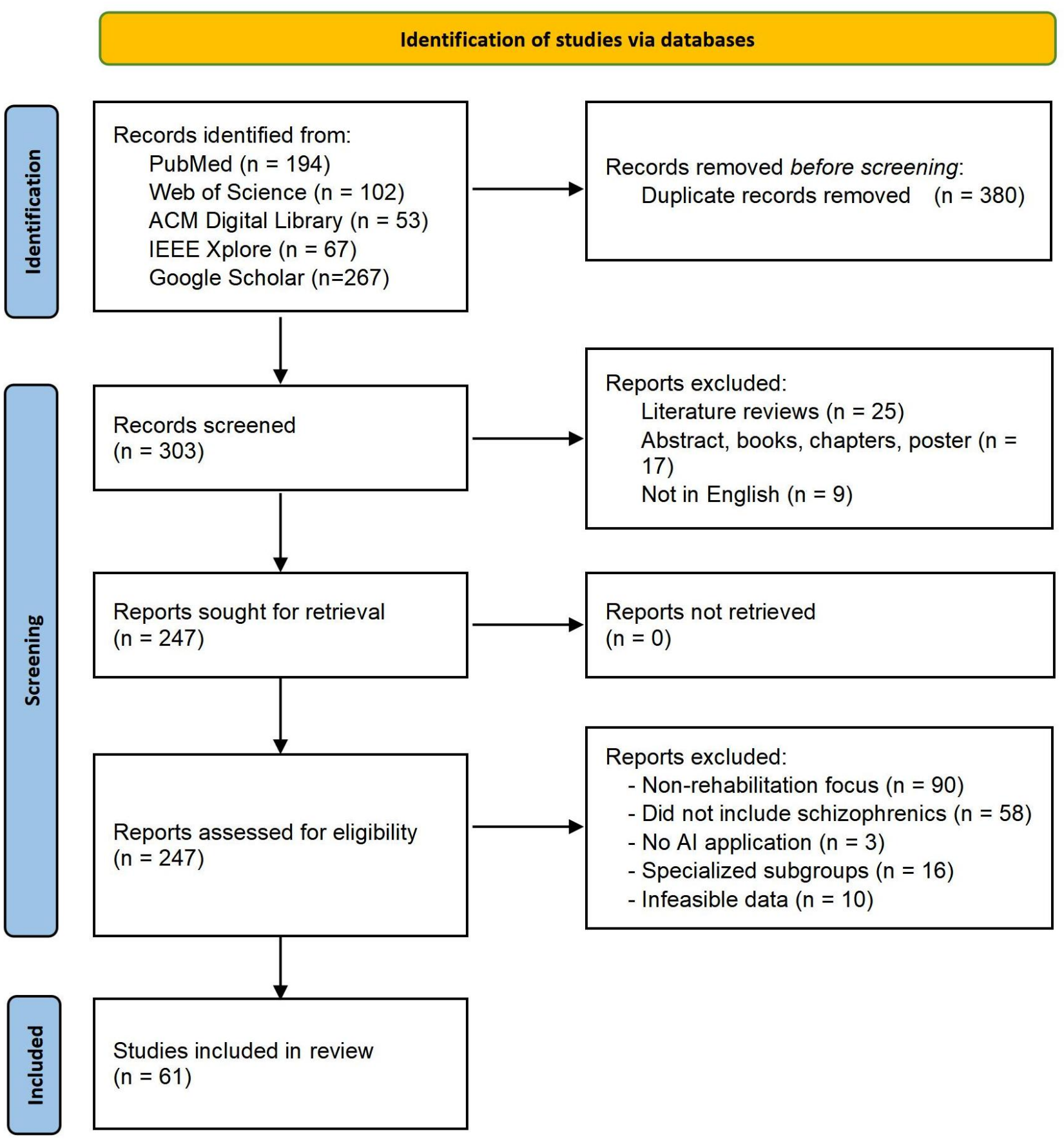

**Figure 2** Flow diagram of the study selection process.

## Data extraction

Two reviewers independently extracted data from eligible studies using Microsoft Excel, employing a pilot test of 20 articles to refine and standardize the procedure, and resolving any discrepancies through discussion. The extracted information included bibliographic details (author, year of publication, and country/region), sample characteristics (schizophrenia subtype, sample size, and disease duration), AI-related elements encompassing data types (e.g., language, behavior, sensors, and medical records), feature engineering methods (if reported), specific algorithmic models (e.g., RF, SVM, and ResNet), performance metrics (e.g., accuracy, AUC, sensitivity, specificity, and F1-score), and approaches to model interpretability. Additionally, we recorded details of rehabilitation management, including the previously described functional domains and the specific functionalities through which AI supported schizophrenia rehabilitation (e.g., relapse prediction, enhancing medication adherence, assessing social functioning, monitoring symptom changes, and providing psychological support). All extracted data were systematically organized and classified by AI model type to facilitate subsequent analyses and discussions. The characteristics of the included studies are summarized in Table 1.

Declaration statements

Data Availability: Data supporting the findings of this study will be made available upon request. Requests should be addressed to the research team at Shanghai Jiao Tong University (hotlz@sjtu.edu.cn).

Code Availability: No custom code or scripts were used in this study.

Acknowledgements: This work was supported by the 2024 Shanghai Jiao Tong University Key Program for Interdisciplinary Research in Medicine and Engineering (Project No. YG2024ZD24). We thank the clinical experts from the Rehabilitation Department of Shanghai Mental Health Center and the nursing staff from several mental rehabilitation management communities in Shanghai for their insightful guidance and assistance. No funding was granted for this study.

Author Contributions: H.Y. conceived and designed the study, performed literature collection and eligibility screening, conducted data extraction, drafted the manuscript, and created the figures. Z.L. assisted in eligibility screening and data extraction. D.Z. participated in the literature collection and eligibility screening. M.F. and F.C. contributed to the figure refinement and manuscript proofreading. All authors reviewed and approved the final version of the manuscript.

Competing Interests: The authors declare no competing interests.

## Reference

1. Stevens, J. R. & Gold, J. M. What is schizophrenia? Behav. Brain Sci. 14, 50–51 (1991). 10.1017/S0140525X00065377.
2. Insel, T. R. Rethinking schizophrenia. Nature 468, 187–193 (2010). 10.1038/nature09552, PubMed: 21068826.
3. McCutcheon, R. A., Reis Marques, T. R. & Howes, O. D. Schizophrenia—an overview. JAMA Psychiatry 77, 201–210 (2020). 10.1001/jamapsychiatry.2019.3360, PubMed: 31664453.
4. WHO. Schizophrenia. https://www.who.int/news-room/fact-sheets/detail/schizophrenia.
5. Solmi, M. et al. Incidence, prevalence, and global burden of schizophrenia – data, with critical appraisal, from the Global Burden of Disease (GBD) 2019. Mol. Psychiatry 28, 5319–5327 (2023). 10.1038/s41380-023-02138-4, PubMed: 37500825.
6. WHO. Comprehensive mental health action plan 2013-2030. https://www.who.int/publications/i/item/9789240031029.
7. Kopelowicz, A., Wallace, C. J. & Liberman, R. P. Psychiatric rehabilitation for schizophrenia. Int. J. Psychol. Psychol. Ther. 3, 283-298 (2003).
8. Morin, L. & Franck, N. Rehabilitation interventions to promote recovery from schizophrenia: a systematic review. Front. Psychiatry 8, 100 (2017). 10.3389/fpsyt.2017.00100, PubMed: 28659832.
9. Lukoff, D., Liberman, R. P. & Nuechterlein, K. H. Symptom monitoring in the rehabilitation of schizophrenic patients. Schizophr. Bull. 12, 578–602 (1986). 10.1093/schbul/12.4.578, PubMed: 3810065.
10. Rössler, W. Psychiatric rehabilitation today: an overview. World Psychiatry 5, 151–157 (2006). PubMed: 17139342.
11. Bellack, A. S., Gold, J. M. & Buchanan, R. W. Cognitive rehabilitation for schizophrenia: problems, prospects, and strategies. Schizophr. Bull. 25, 257-274 (1999). 10.1093/oxfordjournals.schbul.a033377, PubMed: 10416730.
12. Hogarty, G. E. et al. Cognitive enhancement therapy for schizophrenia: effects of a 2-year randomized trial on cognition and behavior. Arch. Gen. Psychiatry 61, 866-876 (2004). 10.1001/archpsyc.61.9.866, PubMed: 15351765.
13. Asher, L., Patel, V. & De Silva, M. J. Community-based psychosocial interventions for people with schizophrenia in low and middle-income countries: systematic review and meta-analysis. BMC Psychiatry 17, 355 (2017). 10.1186/s12888-017-1516-7, PubMed: 29084529.
14. Walker, E. R., Cummings, J. R., Hockenberry, J. M. & Druss, B. G. Insurance status, use of mental health services, and unmet need for mental health care in the United States. Psychiatr. Serv. 66, 578–584 (2015). 10.1176/appi.ps.201400248, PubMed: 25726980.
15. Ben-Zeev, D., Buck, B., Kopelovich, S. & Meller, S. A technology-assisted life of recovery from psychosis. npj Schizophr. 5, 15 (2019). 10.1038/s41537-019-0083-y, PubMed: 31534139.
16. Lattie, E. G., Stiles‐Shields, C. & Graham, A. K. An overview of and recommendations for more accessible digital mental health services. Nat. Rev. Psychol. 1, 87–100 (2022). 10.1038/s44159-021-00003-1, PubMed: 38515434.
17. Bond, R. R. et al. Digital transformation of mental health services. Npj Ment. Health Res. 2, 13 (2023). 10.1038/s44184-023-00033-y, PubMed: 38609479.
18. Haidt, J. & Allen, N. Scrutinizing the effects of digital technology on mental health. Nature 578, 226–227 (2020). 10.1038/d41586-020-00296-x, PubMed: 32042091.
19. Firth, J. & Torous, J. Smartphone apps for schizophrenia: a systematic review. JMIR mHealth uHealth 3, e102 (2015). 10.2196/mhealth.4930, PubMed: 26546039.
20. Schlosser, D. A. et al. Efficacy of PRIME, a mobile app intervention designed to improve motivation in young people with schizophrenia. Schizophr. Bull. 44, 1010-1020 (2018). 10.1093/schbul/sby078, PubMed: 29939367.
21. Almeida, R. F. S. D. et al. Development of weCope, a mobile app for illness self-management in schizophrenia. Arch. Clin. Psychiatry (São Paulo) 46, 1-4 (2019). 10.1590/0101-60830000000182.
22. Fernández-Sotos, P., Fernández-Caballero, A. & Rodriguez-Jimenez, R. Virtual reality for psychosocial remediation in schizophrenia: a systematic review. Eur. J. Psychiatry 34, 1-10 (2020). 10.1016/j.ejpsy.2019.12.003.
23. Park, K. M. et al. A virtual reality application in role-plays of social skills training for schizophrenia: a randomized, controlled trial. Psychiatry Res. 189, 166-172 (2011). 10.1016/j.psychres.2011.04.003, PubMed: 21529970.
24. Krzystanek, M., Borkowski, M., Skałacka, K. & Krysta, K. A telemedicine platform to improve clinical parameters in paranoid schizophrenia patients: results of a one-year randomized study. Schizophr. Res. 204, 389-396 (2019). 10.1016/j.schres.2018.08.016, PubMed: 30154027.
25. Basit, S. A., Mathews, N. & Kunik, M. E. Telemedicine interventions for medication adherence in mental illness: A systematic review. Gen. Hosp. Psychiatry 62, 28-36 (2020). 10.1016/j.genhosppsych.2019.11.004, PubMed: 31775066.

26. Fonseka, L. N. & Woo, B. K. P. Wearables in schizophrenia: update on current and future clinical applications. JMIR mHealth u Health 10, e35600 (2022). 10.2196/35600, PubMed: 35389361.
27. Marsch, L. A. Digital health data-driven approaches to understand human behavior. Neuropsychopharmacology 46, 191-196 (2021). 10.1038/s41386-020-0761-5, PubMed: 32653896.
28. Boonstra, T. W. et al. Using mobile phone sensor technology for mental health research: integrated analysis to identify hidden challenges and potential solutions. J. Med. Internet Res. 20, e10131 (2018). 10.2196/10131, PubMed: 30061092.
29. Arnautovska, U. et al. Efficacy of user self-led and human-supported digital health interventions for people with schizophrenia: A systematic review and meta-analysis. Schizophr. Bull., sbae143 (2024). 10.1093/schbul/sbae143, PubMed: 39340312.
30. Ebert, D. D., Harrer, M., Apolinário-Hagen, J. & Baumeister, H. Digital Interventions for Mental Disorders: Key Features, Efficacy, and Potential for Artificial Intelligence Applications. (ed. Kim, YK) Frontiers in Psychiatry: Artificial Intelligence, Precision Medicine, and Other Paradigm Shifts (583-627), (2019).
31. Russell, S. J. & Norvig, P. Artificial intelligence: a modern approach. (Pearson 2016).
32. Jordan, M. I. & Mitchell, T. M. Machine learning: trends, perspectives, and prospects. Science 349, 255-260 (2015). 10.1126/science.aaa8415, PubMed: 26185243.
33. Freedman, D. A. Statistical Models: Theory and Practice (Cambridge Univ. Press, 2009). 10.1017/CBO9780511815867.
34. Cox, D. R. The regression analysis of binary sequences. J. R. Stat. Soc. B (Methodol.) 20, 215–232 (1958). 10.1111/j.2517-6161.1958.tb00292.x.
35. Quinlan, J. R. Induction of decision trees. Mach. Learn. 1, 81-106 (1986). 10.1007/BF00116251.
36. Cortes, C. & Vapnik, V. Support-vector networks. Mach. Learn. 20, 273-297 (1995). 10.1007/BF00994018.
37. Friedman, J. H. Greedy function approximation: A gradient boosting machine. Ann. Stat. 29, 1189-1232 (2001).
38. Chen, T. & Guestrin, C. XGBoost: A scalable tree boosting system. in Proceedings of the 22nd ACM SIGKDD International Conference on Knowledge Discovery and Data Mining (KDD) (785–794) (ACM, New York, USA, 2016). 10.1145/2939672.2939785.
39. MacQueen, J., (1967). Some methods for classification and analysis of multivariate observations. in Proceedings of the Fifth Berkeley Symposium on Math. Statistics and Probability (281–297).
40. Dempster, A. P., Laird, N. M. & Rubin, D. B. Maximum likelihood from incomplete data via the EM algorithm. J. R. Stat. Soc. B 39, 1-22 (1977). 10.1111/j.2517-6161.1977.tb01600.x.
41. Pearson, K. LIII. On lines and planes of closest fit to systems of points in space. Lond. Edinb. Dublin Philos. Mag. J. Sci. 2, 559-572 (1901). 10.1080/14786440109462720.
42. Goodfellow, I. Bengio, Y Courville, A. Deep Learning, (The MIT Press, 2016).
43. Simonyan, K. & Zisserman, A. Very Deep Convolutional Networks for Large-Scale Image Recognition. ICLR, (2015).
44. He, K., Zhang, X., Ren, S. & Sun, J. Deep residual learning for image recognition. in CVPR (770-778) (IEEE, New York, 2016). 10.1109/CVPR.2016.90.
45. Hochreiter, S. & Schmidhuber, J. Long short-term memory. Neural Comput. 9, 1735-1780 (1997). 10.1162/neco.1997.9.8.1735, PubMed: 9377276.
46. Cho, K. et al. Learning phrase representations using RNN encoder–decoder for statistical machine translation. in EMNLP (1724–1734) (Assoc. for Computational Linguistics, Stroudsburg, PA, USA, 2014). 10.3115/v1/D14-1179.
47. Vaswani, A. et al. Attention is all you need. NeurIPS, 5998-6008 (2017).
48. Mikolov, T., Chen, K., Corrado, G. & Dean, J., (2013). Efficient Estimation of Word Representations in Vector Space. ICLR Workshop.
49. Devlin, J., Chang, M.-W., Lee, K. & Toutanova, K. BERT: pre-training of deep bidirectional transformers for language understanding. NAACL-HLT, 4171–4186 (2019).
50. Brown, T. B. et al. Language models are few-shot learners. NeurIPS, 1877-1901 (2020).
51. Min, B. et al. Recent advances in natural language processing via large pre-trained language models: a survey. ACM Comput. Surv. 56, 1–40 (2023). 10.1145/3605943.
52. King, D. R. et al. An Introduction to Generative Artificial Intelligence in Mental Health Care: Considerations and Guidance. Curr. Psychiatry Rep. 25, 839-846 (2023). 10.1007/s11920-023-01477-x, PubMed: 38032442.
53. Xian, X., Chang, A., Xiang, Y. T. & Liu, M. T. Debate and dilemmas regarding generative AI in mental health care: scoping review. Interact. J. Med. Res. 13, e53672 (2024). 10.2196/53672, PubMed: 39133916.
54. Sutton, R. S. & Barto, A. G. Reinforcement Learning: an Introduction. IEEE Transactions on Neural Networks, 16(1), 285 – 286.

(2005). https://doi.org/10.1109/tnn.2004.842673

55. Arulkumaran, K., Deisenroth, M. P., Brundage, M. & Bharath, A. A. Deep reinforcement learning: A brief survey. IEEE Signal Pro cess. Mag. 34, 26-38 (2017). 10.1109/MSP.2017.2743240.
56. Lillicrap, T. P. et al. Continuous Control with Deep Reinforcement Learning. ICLR, (2016). arXiv (Cornell University). https://arxiv.o rg/pdf/1509.02971.pdf
57. Fujimoto, S., van Hoof, H. & Meger, D. Addressing function approximation error in actor-critic methods. ICML, 1587–1596 TD3 (2018).
58. Fujimoto, S., Meger, D. & Precup, D. (2019). Off-policy deep reinforcement learning without exploration. ICML, 2052–2062. 205 2–2062. http://proceedings.mlr.press/v97/fujimoto19a/fujimoto19a.pdf.
59. Anderson, K. M. et al. Transcriptional and imaging-genetic association of cortical interneurons, brain function, and schizophrenia risk. Nat. Commun. 11, 2889 (2020). 10.1038/s41467-020-16710-x, PubMed: 32514083.
60. Stauffer, E. M., et al.. The genetic relationships between brain structure and schizophrenia. Nature Communications, 14(1). htt ps://doi.org/10.1038/s41467-023-43567-7
61. Kong, L. et al. The network characteristics in schizophrenia with prominent negative symptoms: a multimodal fusion study. Schi zophrenia (Heidelb) 10, 10 (2024). 10.1038/s41537-023-00408-2, PubMed: 38233433.
62. Karaglani, M. et al. A novel blood-based epigenetic biosignature in first-episode schizophrenia patients through automated mach ine learning. Translational Psychiatry 14, (2024).
63. Bracher-Smith, M. et al. Machine learning for prediction of schizophrenia using genetic and demographic factors in the UK Bio bank. Schizophr. Res. 246, 156–164 (2022). 10.1016/j.schres.2022.06.006, PubMed: 35779327.
64. Starke, G., De Clercq, E., Borgwardt, S. & Elger, B. S. Computing schizophrenia: ethical challenges for machine learning in psych iatry. Psychol. Med. 51, 2515–2521 (2021). 10.1017/S0033291720001683, PubMed: 32536358.
65. Yoo, D. W. et al. Patient perspectives on AI-driven predictions of schizophrenia relapses: understanding concerns and opportunit ies for self-care and treatment. Proc. SIGCHI Conf. Hum. Factor. Comput. Syst. 2024, 702 (2024). 10.1145/3613904.3642369, Pu bMed: 38894725.
66. Gashkarimov, V. R., Sultanova, R. I., Efremov, I. S. & Asadullin, A. R. Machine learning techniques in diagnostics and prediction of the clinical features of schizophrenia: a narrative review. Consortium Psychiatricum 4, 43-53 (2023). 10.17816/CP11030, Pub Med: 38249535.
67. Sharma, M., Patel, R. K., Garg, A., SanTan, R. & Acharya, U. R. Automated detection of schizophrenia using deep learning: a r eview for the last decade. Physiol. Meas. 44, 03TR01 (2023). 10.1088/1361-6579/acb24d, PubMed: 36630717.
68. Thakkar, A., Gupta, A. & De Sousa, A. Artificial intelligence in positive mental health: a narrative review. Front. Digit. Health 6, 1280235 (2024). 10.3389/fdgth.2024.1280235, PubMed: 38562663.
69. Lee, E. E. et al. Artificial intelligence for mental health care: clinical applications, barriers, facilitators, and artificial wisdom. Biol. Psychiatry Cogn. Neurosci. Neuroimaging 6, 856-864 (2021). 10.1016/j.bpsc.2021.02.001, PubMed: 33571718.
70. Howes, C., Purver, M., McCabe, R., Healey, P. G. T. & Lavelle, M., (2012). Predicting adherence to treatment for schizophrenia from dialogue transcripts. in in Proceedings of the 13th Annual Meeting of the Special Interest Group on Discourse and Dialog ue (79-83). https://www.eecs.qmul.ac.uk/~mpurver/papers/howes-et-al12sigdial.pdf.
71. Osipov, M., Behzadi, Y., Kane, J. M., Petrides, G. & Clifford, G. D. Objective identification and analysis of physiological and beh avioral signs of schizophrenia. J. Ment. Health 24, 276–282 (2015). 10.3109/09638237.2015.1019048, PubMed: 26193048.
72. Wang, R. et al. CrossCheck: toward passive sensing and detection of mental health changes in people with schizophrenia. in in Proceedings of the 2016 ACM International Joint Conference on Pervasive and Ubiquitous Computing, ACM, New York, USA 8 86-897 (2016). 10.1145/2971648.2971740.
73. Wang, R. et al. Predicting symptom trajectories of schizophrenia using mobile sensing. Proc. ACM Interact. Mob. Wearable Ubiq uitous Technol. 1, 1–24 (2017). 10.1145/3130976.
74. Birnbaum, M. L., Ernala, S. K., Rizvi, A. F., De Choudhury, M. & Kane, J. M. A collaborative approach to identifying social medi a markers of schizophrenia by employing machine learning and clinical appraisals. J. Med. Internet Res./Journal of Medical Inte rnet Research 19, e289 (2017). 10.2196/jmir.7956, PubMed: 28807891.
75. Fond, G. et al. Machine learning for predicting psychotic relapse at 2 years in schizophrenia in the national FACE-SZ cohort. Pr og. Neuropsychopharmacol. Biol. Psychiatry 92, 8–18 (2019). 10.1016/j.pnpbp.2018.12.005, PubMed: 30552914.
76. Brodey, B. B. et al. The Early Psychosis Screener for Internet (EPSI)-SR: predicting 12 month psychotic conversion using machin

e learning. Schizophr. Res. 208, 390–396 (2019). 10.1016/j.schres.2019.01.015, PubMed: 30777603.

77. Hulme, W. J. et al. Cluster hidden markov models: an application to ecological momentary assessment of schizophrenia. in in 32nd International Symposium on Computer-Based Medical Systems (CBMS) (99-103) (IEEE, 2019; 2019). 10.1109/CBMS.2019.00030.

78. Cohen, A. S. et al. Using machine learning of computerized vocal expression to measure blunted vocal affect and alogia. npj Schizophr. 6, 26 (2020). 10.1038/s41537-020-00115-2, PubMed: 32978400.

79. Wu, C. S. et al. Development and validation of a machine learning Individualized treatment rule in First-Episode schizophrenia. JAMA Netw. Open 3, e1921660 (2020). 10.1001/jamanetworkopen.2019.21660, PubMed: 32083693.

80. Wang, K. Z. et al. Prediction of physical violence in schizophrenia with machine learning algorithms. Psychiatry Res. 289, 112960 (2020). 10.1016/j.psychres.2020.112960, PubMed: 32361562.

81. Birnbaum, M. L. et al. Utilizing Machine Learning on Internet Search Activity to Support the Diagnostic Process and Relapse Detection in Young Individuals With Early Psychosis: Feasibility Study. JMIR Ment. Health 7, e19348 (2020). 10.2196/19348, PubMed: 32870161.

82. Arevian, A. C. et al. Clinical state tracking in serious mental illness through computational analysis of speech. PLOS One 15, e0225695 (2020). 10.1371/journal.pone.0225695, PubMed: 31940347.

83. Tseng, V. W. S. et al. Using behavioral rhythms and multi-task learning to predict fine-grained symptoms of schizophrenia. Sci. Rep. 10, 15100 (2020). 10.1038/s41598-020-71689-1, PubMed: 32934246.

84. Wang, W. et al. Social sensing: Assessing social functioning of patients living with schizophrenia using mobile phone sensing. Proceedings of the 2020 CHI Conference on Human Factors in Computing Systems. 2020. 1–15 (2020) doi:10.1145/3313831.3376855.

85. Wang, X. et al. HOPES: an integrative digital phenotyping platform for data collection, monitoring, and machine learning. J. Med. Internet Res. 23, e23984 (2021). 10.2196/23984, PubMed: 33720028.

86. Banerjee, S., Lió, P., Jones, P. B. & Cardinal, R. N. A class-contrastive human-interpretable machine learning approach to predict mortality in severe mental illness. npj Schizophr. 7, 60 (2021). 10.1038/s41537-021-00191-y, PubMed: 34880262.

87. De Boer, J. N. et al. Acoustic speech markers for schizophrenia-spectrum disorders: a diagnostic and symptom-recognition tool. Psychol. Med. 53, 1302–1312 (2023). 10.1017/S0033291721002804, PubMed: 34344490.

88. Ziv, I. et al. Morphological characteristics of spoken language in schizophrenia patients – an exploratory study. Scand. J. Psychol. 63, 91–99 (2022). 10.1111/sjop.12790, PubMed: 34813111.

89. Lin, E., Lin, C. H. & Lane, H. Y. Applying a bagging ensemble machine learning approach to predict functional outcome of schizophrenia with clinical symptoms and cognitive functions. Sci. Rep. 11, 6922 (2021). 10.1038/s41598-021-86382-0, PubMed: 33767310.

90. Jeon, G. et al. Evaluation of the correlation between gaze avoidance and schizophrenia psychopathology with deep learning-based emotional recognition. Asian J. Psychiatry 68, 102974 (2022). 10.1016/j.ajp.2021.102974, PubMed: 34974374.

91. Parola, A., Gabbatore, I., Berardinelli, L., Salvini, R. & Bosco, F. M. Multimodal assessment of communicative-pragmatic features in schizophrenia: a machine learning approach. npj Schizophr. 7, 28 (2021). 10.1038/s41537-021-00153-4, PubMed: 34031425.

92. Rashid, N. A. et al. Evaluating the utility of digital phenotyping to predict health outcomes in schizophrenia: protocol for the HOPE-S observational study. BMJ Open 11, e046552 (2021). 10.1136/bmjopen-2020-046552, PubMed: 34670760

93. Podichetty, J. T. et al. Application of machine learning to predict reduction in total PANSS score and enrich enrollment in schizophrenia clinical trials. Clin. Transl. Sci. 14, 1864–1874 (2021). 10.1111/cts.13035, PubMed: 33939284.

94. Soldatos, R. F. et al. Prediction of early symptom remission in two independent samples of first-episode psychosis patients using machine learning. Schizophr. Bull. 48, 122-133 (2022). 10.1093/schbul/sbab107, PubMed: 34535800.

95. Badal, V. D. et al. Computational methods for integrative evaluation of confidence, accuracy, and reaction time in facial affect recognition in schizophrenia. Schizophr. Res. Cogn. 25, 100196 (2021). 10.1016/j.scog.2021.100196, PubMed: 33996517.

96. Depp, C. A. et al. Ecological momentary facial emotion recognition in psychotic disorders. Psychol. Med. 52, 2531–2539 (2022). 10.1017/S0033291720004419, PubMed: 33431072.

97. Narkhede, S. M. et al. Machine learning identifies digital phenotyping measures most relevant to negative symptoms in psychotic disorders: implications for clinical trials. Schizophr. Bull. 48, 425–436 (2022). 10.1093/schbul/sbab134, PubMed: 34915570.

98. Kalinich, M. et al. Applying machine learning to smartphone based cognitive and sleep assessments in schizophrenia. Schizophr. Res. Cogn. 27, 100216 (2022). 10.1016/j.scog.2021.100216, PubMed: 34934638.

99. Price, G. D. et al. An unsupervised machine learning approach using passive movement data to understand depression and schizophrenia. J. Affect. Disord. (Print) 316, 132–139 (2022). 10.1016/j.jad.2022.08.013, PubMed: 35964770.

100. Adler, D. A., Wang, F., Mohr, D. C. & Choudhury, T. Machine learning for passive mental health symptom prediction: generalization across different longitudinal mobile sensing studies. PLOS One 17, e0266516 (2022). 10.1371/journal.pone.0266516, PubMed: 35476787.

101. Jeon, S. M., Cho, J., Lee, D. Y. & Kwon, J. W. Comparison of prediction methods for treatment continuation of antipsychotics in children and adolescents with schizophrenia. Evid. Based Ment. Health 25, e26–e33 (2022). 10.1136/ebmental-2021-300404, PubMed: 35418448.

102. Zhu, X. et al. Integrating machine learning with electronic health record data to facilitate detection of prolactin level and pharmacovigilance signals in olanzapine-treated patients. Front. Endocrinol. 13, 1011492 (2022). 10.3389/fendo.2022.1011492, PubMed: 36313772.

103. Richter, V. et al., (2022). Towards multimodal dialog-based speech & facial biomarkers of schizophrenia. 10.1145/3536220.3558075.

104. Chen, H. H. et al. Efficacy of a smartphone app in enhancing medication adherence and accuracy in individuals with schizophrenia during the COVID-19 pandemic: randomized controlled trial. JMIR Ment. Health 10, e50806 (2023). 10.2196/50806, PubMed: 38096017.

105. Hudon, A., Beaudoin, M., Phraxayavong, K., Potvin, S. & Dumais, A. Unsupervised machine learning driven analysis of verbatims of Treatment-Resistant schizophrenia patients having followed Avatar therapy. J. Pers. Med. 13, 801 (2023). 10.3390/jpm13050801, PubMed: 37240971.

106. Miley, K. et al. Causal pathways to social and occupational functioning in the first episode of schizophrenia: uncovering unmet treatment needs. Psychol. Med. 53, 2041–2049 (2023). 10.1017/S0033291721003780, PubMed: 37310333.

107. Wong, T. Y. et al. Development of an individualized risk calculator of treatment resistance in patients with first-episode psychosis (TRipCal) using automated machine learning: a 12-year follow-up study with clozapine prescription as a proxy indicator. Transl. Psychiatry 14, 50 (2024). 10.1038/s41398-024-02754-w, PubMed: 38253484.

108. Difrancesco, S. et al. Out-of-Home activity recognition from GPS data in schizophrenic patients. in 29th International Symposium on Computer-based Medical Systems (CBMS) (324-328) (IEEE, 2016; 2016). 10.1109/CBMS.2016.54.

109. Amoretti, S. et al. Identifying clinical clusters with distinct trajectories in first-episode psychosis through an unsupervised machine learning technique. Euro. Neuropsychopharmacol. 47, 112–129 (2021). 10.1016/j.euroneuro.2021.01.095.

110. Rodríguez-Ruiz, J. G. et al. Classification of depressive and schizophrenic episodes using Night-Time Motor Activity Signal. Healthcare (Basel) 10, 1256 (2022). 10.3390/healthcare10071256, PubMed: 35885784.

111. Zhou, J., Lamichhane, B., Ben‐Zeev, D., Campbell, A. T. & Sano, A. Predicting psychotic relapse in schizophrenia with mobile sensor data: routine cluster analysis. JMIR mHealth uHealth 10, e31006 (2022). 10.2196/31006, PubMed: 35404256.

112. Holmlund, T. B. et al. Applying speech technologies to assess verbal memory in patients with serious mental illness. npj Digit. Med. 3, 33 (2020). 10.1038/s41746-020-0241-7, PubMed: 32195368.

113. Xu, W. et al. Fully automated detection of formal thought disorder with Time-series Augmented Representations for Detection of Incoherent Speech (TARDIS). J. Biomed. Inform. (Print) 126, 103998 (2022). 10.1016/j.jbi.2022.103998, PubMed: 35063668.

114. Lee, J., Kopelovich, S. L., Cheng, S. C. & Si, D., Psychosis iREACH: reach for psychosis treatment using artificial intelligence 2022 IEEE International Conference on Bioinformatics and Biomedicine (BIBM), Las Vegas, NV, USA 2913-2920 (2022). 10.1109/BIBM55620.2022.9995430.

115. Ciampelli, S., Voppel, A. E., De Boer, J. N., Koops, S. & Sommer, I. E. C. Combining automatic speech recognition with semantic natural language processing in schizophrenia. Psychiatry Res. 325, 115252 (2023). 10.1016/j.psychres.2023.115252, PubMed: 37236098.

116. Jeong, L. et al. Exploring the use of natural language processing for objective assessment of disorganized speech in schizophrenia. Psychiatr. Res. Clin. Pract. 5, 84–92 (2023). 10.1176/appi.prcp.20230003, PubMed: 37711756.

117. Chan, C. C. et al. Emergence of language related to self-experience and agency in autobiographical narratives of individuals with schizophrenia. Schizophr. Bull. 49, 444–453 (2023). 10.1093/schbul/sbac126, PubMed: 36184074.

118. Just, S. A. et al. Validation of natural language processing methods capturing semantic incoherence in the speech of patients with non-affective psychosis. Front. Psychiatry 14, 1208856 (2023). 10.3389/fpsyt.2023.1208856, PubMed: 37564246.

119. Shibata, Y. et al. Estimation of subjective quality of life in schizophrenic patients using speech features. Front. Rehabil. Sci. 4,

1121034 (2023). 10.3389/fresc.2023.1121034, PubMed: 36968213.

120. Wang, W. et al. The power of speech in the wild: discriminative power of daily voice diaries in understanding auditory verbal hallucinations using deep learning. Proc. ACM Interact. Mob. Wearable Ubiquitous Technol. 7, 1-29 (2023). 10.1145/3610890.
121. Bain, E. E. et al. Use of a novel artificial intelligence platform on mobile devices to assess dosing compliance in a phase 2 cli nical trial in subjects with schizophrenia. JMIR mHealth uHealth 5, e18 (2017). 10.2196/mhealth.7030, PubMed: 28223265.
122. Shen, H. et al. Color painting predicts clinical symptoms in chronic schizophrenia patients via deep learning. BMC Psychiatry 21, 522 (2021). 10.1186/s12888-021-03452-3, PubMed: 34686178.
123. Cotes, R. O. et al. Multimodal assessment of schizophrenia and depression utilizing video, acoustic, locomotor, electroencephalo graphic, and heart rate technology: protocol for an observational study. JMIR Res. Protoc. 11, e36417 (2022). 10.2196/36417, P ubMed: 35830230.
124. Martin, L., Stein, K., Kubera, K. M., Troje, N. F. & Fuchs, T. Movement markers of schizophrenia: a detailed analysis of patients' gait patterns. Eur. Arch. Psychiatry Clin. Neurosci. 272, 1347–1364 (2022). 10.1007/s00406-022-01402-y, PubMed: 35362775.
125. Koppe, G., Gülöksüz, S., Reininghaus, U. & Durstewitz, D. Recurrent neural networks in mobile sampling and intervention. Schiz ophr. Bull. 45, 272–276 (2019). 10.1093/schbul/sby171, PubMed: 30496527.
126. Umbricht, Cheng, D., W., Lipsmeier, F., Bamdadian, A. & Lindemann, M. Deep Learning-Based Human Activity Recognition for co ntinuous activity and Gesture monitoring for schizophrenia patients with negative symptoms. Front. Psychiatry 11, 574375 (202 0). 10.3389/fpsyt.2020.574375, PubMed: 33192706.
127. Adler, D. A. et al. Predicting early warning signs of psychotic relapse from passive sensing data: an approach using Encoder-De coder neural networks. JMIR mHealth uHealth 8, e19962 (2020). 10.2196/19962, PubMed: 32865506.
128. Nguyen, D. K., Chan, C. L., Li, A. A., Phan, D. V. & Lan, C. H. Decision support system for the differentiation of schizophrenia and mood disorders using multiple deep learning models on wearable devices data. Health Inform. J. 28, 14604582221137537 (2022). 10.1177/14604582221137537, PubMed: 36317536.
129. 139. Zlatintsi, A. et al. E-prevention: advanced support system for monitoring and relapse prevention in patients with psychotic disorders analyzing long-term multimodal data from wearables and video captures. Sensors (Basel) 22, 7544 (2022a). 10.3390/s 22197544, PubMed: 36236643.
130. Lin, B., Cecchi, G. A. & Bouneffouf, D., (2023). Psychotherapy AI companion with reinforcement learning recommendations and interpretable policy dynamics. in in Companion Proc. ACM Web Conference (932-939). 10.1145/3543873.3587623.
131. Saboori Amleshi, R. S. et al. Predictive utility of artificial intelligence on schizophrenia treatment outcomes: A systematic review and meta-analysis. Neurosci. Biobehav. Rev. 169, 105968 (2024). 10.1016/j.neubiorev.2024.105968, PubMed: 39643220.
132. Zhou, S., Zhao, J. & Zhang, L. Application of artificial intelligence on psychological interventions and diagnosis: an overview. Fro nt. Psychiatry 13, 811665 (2022). 10.3389/fpsyt.2022.811665, PubMed: 35370846.
133. Cohen, A. et al. Relapse prediction in schizophrenia with smartphone digital phenotyping during COVID-19: a prospective, three -site, two-country, longitudinal study. Schizophrenia (Heidelb) 9, 6 (2023). 10.1038/s41537-023-00332-5, PubMed: 36707524.
134. Yan, W. J., Ruan, Q. N. & Jiang, K. Challenges for artificial intelligence in recognizing mental disorders. Diagnostics (Basel) 13, 2 (2022). 10.3390/diagnostics13010002, PubMed: 36611294.
135. Chancellor, S. & De Choudhury, M. Methods in predictive techniques for mental health status on social media: a critical review. npj Digit. Med. 3, 43 (2020). 10.1038/s41746-020-0233-7, PubMed: 32219184.
136. Fonseka, L. N. & Woo, B. K. P. Social media and schizophrenia: an update on clinical applications. World J. Psychiatry 12, 897–903 (2022). 10.5498/wjp.v12.i7.897, PubMed: 36051600.
137. Betton, V. et al. The role of social media in reducing stigma and discrimination. Br. J. Psychiatry 206, 443–444 (2015). 10.1192 /bjp.bp.114.152835, PubMed: 26034176.
138. 148. Zomick, J., Levitan, S. I. & Serper, M. R. Linguistic analysis of schizophrenia in. in in Proceedings of the Sixth Workshop o n Computational Linguistics and Clinical Psychology (74-83) (Assoc. for Computational Linguistics, Stroudsburg, PA, USA, 2019). 1 0.18653/v1/W19-3009.
139. Passerello, G. L., Hazelwood, J. E. & Lawrie, S. M. Using Twitter to assess attitudes to schizophrenia and psychosis. BJPsych Bul l. 43, 158–166 (2019). 10.1192/bjb.2018.115, PubMed: 30784393.
140. Liang, H., Sun, X., Sun, Y. & Gao, Y. Text feature extraction based on deep learning: a review. EURASIP J. Wirel. Commun. Net w. 2017, 211 (2017). 10.1186/s13638-017-0993-1, PubMed: 29263717.
141. Garg, D., Verma, G. K. & Singh, A. K. [RETRACTED ARTICLE] A review of deep learning based methods for affect analysis using

physiological signals. Multimedia Tool. Appl. 82, 26089-26134 (2023). 10.1007/s11042-023-14354-9.

142. Olson, G. M., Damme, K. S. F., Cowan, H. R., Alliende, L. M. & Mittal, V. A. Emotional tone in clinical high risk for psychosis: novel insights from a natural language analysis approach. Front. Psychiatry 15, 1389597 (2024). 10.3389/fpsyt.2024.1389597, PubMed: 38803678.
143. Bizzego, A., Gabrieli, G. & Esposito, G. Deep neural networks and transfer learning on a multivariate physiological signal Dataset. Bioengineering (Basel) 8, 35 (2021). 10.3390/bioengineering8030035, PubMed: 33800842.
144. Li, H. Deep learning for natural language processing: advantages and challenges. Natl Sci. Rev. 5, 24-26 (2018). 10.1093/nsr/nwx110.
145. Dzieżyc, M., Gjoreski, M., Kazienko, P., Saganowski, S. & Gams, M. Can we ditch feature engineering? end-to-end deep learning for affect recognition from physiological sensor data. Sensors (Basel) 20, 6535 (2020). 10.3390/s20226535, PubMed: 33207564.
146. Alqahtani, A. et al. An efficient approach for textual data classification using deep learning. Front. Comp. Neurosci. 16, 992296 (2022). 10.3389/fncom.2022.992296, PubMed: 36185709.
147. Arik, S. Ö. & Pfister, T. Tabnet: attentive interpretable tabular learning. AAAI 35, 6679–6687 (2021). 10.1609/aaai.v35i8.16826.
148. Manríquez Roa, T. & Biller‐Andorno, N. Black box algorithms in mental health apps: an ethical reflection. Bioethics 37, 790-797 (2023). 10.1111/bioe.13215, PubMed: 37503823.
149. Graham, S. et al. Artificial intelligence for mental health and mental illnesses: an overview. Curr. Psychiatry Rep. 21, 116 (2019). 10.1007/s11920-019-1094-0, PubMed: 31701320.
150. Achtibat, R. et al. From attribution maps to human-understandable explanations through concept relevance propagation. Nat. Mach. Intell. 5, 1006-1019 (2023). 10.1038/s42256-023-00711-8.
151. Saporta, A. et al. Benchmarking saliency methods for chest X-ray interpretation. Nat. Mach. Intell. 4, 867-878 (2022). 10.1038/s42256-022-00536-x.
152. Hosain, M. T., Jim, J. R., Mridha, M. F. & Kabir, M. M. Explainable AI approaches in deep learning: advancements, applications and challenges. Comput. Electr. Eng. 117, 109246 (2024). 10.1016/j.compeleceng.2024.109246.
153. Thirunavukarasu, A. J. & OʹLogbon, J. The potential and perils of generative artificial intelligence in psychiatry and psychology. Nat. Mental Health 2, 745–746 (2024). 10.1038/s44220-024-00257-7.
154. Perlis, R. H., Goldberg, J. F., Ostacher, M. J. & Schneck, C. D. Clinical decision support for bipolar depression using large language models. Neuropsychopharmacology 49, 1412–1416 (2024). 10.1038/s41386-024-01841-2, PubMed: 38480911.
155. Kim, J. et al. Large language models outperform mental and medical health care professionals in identifying obsessive-compulsive disorder. npj Digit. Med. 7, 193 (2024). 10.1038/s41746-024-01181-x, PubMed: 39030292.
156. Levkovich, I. & Elyoseph, Z. Identifying depression and its determinants upon initiating treatment: ChatGPT versus primary care physicians. Fam. Med. Community Health 11, e002391 (2023). 10.1136/fmch-2023-002391, PubMed: 37844967.
157. Spallek, S., Birrell, L., Kershaw, S., Devine, E. K. & Thornton, L. Can we use ChatGPT for mental health and substance use education? Examining its quality and potential harms. JMIR Med. Educ. 9, e51243 (2023). 10.2196/51243, PubMed: 38032714.
158. Open, A. I. Advanced Voice Mode released [Community post] (OpenAI Community, 2024). https://community.openai.com/t/advanced-voice-mode-released-09252024/956738.
159. Han, T. et al. Medical large language models are susceptible to targeted misinformation attacks. npj Digit. Med. 7, 288 (2024). 10.1038/s41746-024-01282-7, PubMed: 39443664.
160. Obradovich, N. et al. Opportunities and risks of large language models in psychiatry. NPP Digit. Psychiatry Neurosci. 2, 8 (2024). 10.1038/s44277-024-00010-z, PubMed: 39554888.
161. European Parliament. EU AI Act: first regulation on artificial intelligence (European Parliament, 2023). https://www.europarl.europa.eu/topics/en/article/20230601STO93804/eu-ai-act-first-regulation-on-artificial-intelligence.
162. Chung, N. C., Dyer, G. & Brocki, L. (2023). Challenges of Large Language Models for Mental Health Counseling. arXiv Preprint ArXiv:2311.13857. https://doi.org/10.48550/arxiv.2311.13857
163. Ma, Y. et al. Integrating large language models in mental health practice: a qualitative descriptive study based on expert interviews. Front. Public Health 12, 1475867 (2024). 10.3389/fpubh.2024.1475867, PubMed: 39559378.
164. Dewhirst, A., Laugharne, R. & Shankar, R. Therapeutic use of serious games in mental health: scoping review. BJPsych Open 8, e37 (2022). 10.1192/bjo.2022.4, PubMed: 35105418.
165. Nilsen, P. et al. Accelerating the impact of artificial intelligence in mental healthcare through implementation science. Implement. Res. Pract. 3, 26334895221112033 (2022). 10.1177/26334895221112033, PubMed: 37091110.

166. Olawade, D. B. et al. Enhancing mental health with Artificial Intelligence: current trends and future prospects. J. Med. Surg. Pu blic Health 3, 100099 (2024). 10.1016/j.glmedi.2024.100099.
167. Cross, S. et al. Use of AI in mental health care: community and mental health professionals survey. JMIR Ment. Health 11, e6 0589 (2024). 10.2196/60589, PubMed: 39392869.
168. Tricco, A. C. et al. PRISMA Extension for Scoping Reviews (PRISMA-SCR): checklist and explanation. Ann. Intern. Med. 169, 467–473 (2018). 10.7326/M18-0850, PubMed: 30178033.
169. Krizhevsky, A., Sutskever, I. & Hinton, G. E. ImageNet classification with deep convolutional neural networks. Commun. ACM 60, 84–90 (2017). 10.1145/3065386.
170. LeCun, Y., Bengio, Y. & Hinton, G. E. Deep learning. Nature 521, 436–444 (2015). 10.1038/nature14539, PubMed: 26017442.
171. Chukka, A. et al. Digital interventions for relapse prevention, illness self-management, and Health Promotion in schizophrenia: r ecent advances, continued challenges, and future opportunities. Curr. Treat. Options Psychiatry 10, 346-371 (2023). 10.1007/s40 501-023-00309-2.
172. Chivilgina, O., Elger, B. S. & Jotterand, F. Digital technologies for schizophrenia management: A descriptive review. Sci. Eng. Ethi cs 27, 25 (2021). 10.1007/s11948-021-00302-z, PubMed: 33835287.
173. Murphy, S. M., Irving, C. B., Adams, C. E. & Waqar, M. Crisis intervention for people with severe mental illnesses. Cochrane D atabase Syst. Rev. 2015, CD001087 (2015). 10.1002/14651858.CD001087.pub5, PubMed: 26633650.
174. Carr, N. Rehabilitation and recovery from schizophrenia. MentalHealth.com, (2023). https://www.mentalhealth.com/library/rehabilit ation-and-recovery-from-schizophrenia.
175. Rus-Calafell, M, et al. A brief cognitive–behavioural social skills training for stabilised outpatients with schizophrenia: A prelimin ary study. Schizophrenia research 143.2-3 (2013): 327-336. https://doi.org/10.1016/j.schres.2012.11.014
176. Lean, M. et al. Self-management interventions for people with severe mental illness: systematic review and meta-analysis. Br. J. Psychiatry 214, 260-268 (2019). 10.1192/bjp.2019.54, PubMed: 30898177.
177. Chien, W. T., Leung, S. F., Yeung, F. K. & Wong, W. K. Current approaches to treatments for schizophrenia spectrum disorders, part II: psychosocial interventions and patient-focused perspectives in psychiatric care. Neuropsychiatr. Dis. Treat. 9, 1463-1481 (2013). 10.2147/NDT.S49263, PubMed: 24109184.

Figure Legends

**Figure 1.** Trends and Applications of AI in Studies.

This figure is composed of two panels. (a) A bar chart (purple bars) illustrating the number of relevant articles published each year from 2012 to 2024, with a black solid line indicating the overall publication trend. Data for 2024 only includes publications from January 2024. (b) A horizontal bar chart illustrates the frequency distribution of AI techniques, including supervised machine learning (SML), unsupervised machine learning (UML), deep learning (DL), and reinforcement learning (RL). Notably, a single study may employ multiple methods. SML appears most frequently (51 studies), followed by DL (14), UML (7), and RL (3).

**Figure 2.** Flow diagram of the study selection process.

This flowchart details the systematic screening and selection of studies in our review, including the number of record identified, screened, assessed for eligibility, and finally included, along with reasons for exclusion. This transparent and reproducible process follows the PRESMA guidelines.